%% file: aaai25.tex
\documentclass[letterpaper]{article} 
\usepackage{aaai25}
\usepackage{times}  
\usepackage{helvet}  
\usepackage{courier}  
\usepackage[hyphens]{url}  
\usepackage{graphicx} 
\usepackage{natbib}  
\usepackage{caption} 
\usepackage{algorithm}
\usepackage{amsfonts}       
\usepackage{nicefrac}       
\usepackage{microtype}      
\usepackage{amsmath, amssymb, overpic, textpos}
\usepackage{graphicx}
\usepackage{multirow}
\usepackage{colortbl}
\usepackage{tabulary}
\usepackage{algorithmicx}
\usepackage{algpseudocode}
\usepackage{listings}
\usepackage{multicol}
\usepackage{xspace}
\usepackage{graphbox}
\usepackage{arydshln}
\usepackage{algorithm}
\usepackage{adjustbox}
\usepackage{kotex}
\usepackage{caption}
\usepackage{subcaption}
\usepackage{setspace}
\usepackage{makecell}
\usepackage{booktabs}
\usepackage{color}
\usepackage{xcolor}
\definecolor{cvprblue}{rgb}{0.21,0.49,0.74}
\usepackage[pagebackref,breaklinks,colorlinks,citecolor=cvprblue]{hyperref}

\usepackage[capitalize]{cleveref}
\usepackage{pifont}
\usepackage{newfloat}
\usepackage{listings}
\DeclareCaptionStyle{ruled}{labelfont=normalfont,labelsep=colon,strut=off} 

\floatstyle{ruled}
\newfloat{listing}{tb}{lst}{}
\floatname{listing}{Listing}

\crefname{section}{Sec.}{Secs.}
\Crefname{section}{Section}{Sections}
\Crefname{table}{Table}{Tables}
\crefname{table}{Tab.}{Tabs.}

\makeatletter
\DeclareRobustCommand\onedot{\futurelet\@let@token\@onedot}
\def\@onedot{\ifx\@let@token.\else.\null\fi\xspace}
 
\def\ie{\emph{i.e}\onedot} 
 
 \def\vs{\emph{vs}\onedot}

 \def\etal{\emph{et al}\onedot}
\makeatother
\newlength\savewidth\newcommand\shline{\noalign{\global\savewidth\arrayrulewidth
  \global\arrayrulewidth 1pt}\hline\noalign{\global\arrayrulewidth\savewidth}}
\newcommand{\tablestyle}[2]{\setlength{\tabcolsep}{#1}\renewcommand{\arraystretch}{#2}\centering\footnotesize}
\newcolumntype{x}[1]{>{\centering\arraybackslash}p{#1pt}}
\newcolumntype{y}[1]{>{\raggedright\arraybackslash}p{#1pt}}
\newcolumntype{z}[1]{>{\raggedleft\arraybackslash}p{#1pt}}

\newlength\secmargin
\newlength\subsecmargin
\newlength\subsubsecmargin
\newlength\paramargin
\newlength\abovetabcapmargin
\newlength\belowtabcapmargin
\newlength\abovefigcapmargin
\newlength\belowfigcapmargin
\definecolor{Gray}{gray}{0.9}

\definecolor{Green}{rgb}{0.2, 0.7, 0.1}
\definecolor{prompt_blue}{HTML}{1f78b4}
\definecolor{prompt_red}{HTML}{d45c43}
\definecolor{plus}{HTML}{0071bc}
\definecolor{minus}{RGB}{153,10,10}

\newcommand{\up}{\bf \fontsize{10}{42} \color{plus}{$\uparrow$}}
\newcommand{\down}{\bf \fontsize{10}{42}\selectfont {$\downarrow$}}

\input{math_commands.tex}

\title{Diffusion Model Patching via Mixture-of-Prompts}
\author {
    Seokil Ham\textsuperscript{\rm 1}\equalcontrib,
    Sangmin Woo\textsuperscript{\rm 1}\equalcontrib,
    Jin-Young Kim\textsuperscript{\rm 2},
    Hyojun Go\textsuperscript{\rm 2},
    Byeongjun Park\textsuperscript{\rm 1},
    Changick Kim\textsuperscript{\rm 1},
}
\affiliations {
    \textsuperscript{\rm 1}KAIST \quad \textsuperscript{\rm 2}Twelve Labs\\
    \textsuperscript{\rm 1}\{gkatjrdlf, smwoo95, pbj3810, changick\}@kaist.ac.kr \quad \textsuperscript{\rm 2}\{seago0828, gohyojun15\}@gmail.com
}

\begin{document}
\maketitle
\input{contents/00abstract}

\begin{links}
    \link{Project Page}{https://sangminwoo.github.io/DMP/}
\end{links}

\input{contents/01introduction}

\input{contents/02related_work}
\input{contents/03preliminaries}
\input{contents/04approach}

\input{contents/05experiments}
\input{contents/06conclusion}
\bibliography{aaai25}
\input{contents/08appendix}
\end{document}

%% file: math_commands.tex
\usepackage{amsmath,amsfonts,bm}
\usepackage{mathtools}

\def\eqref#1{equation~\ref{#1}}

\def\1{\bm{1}}

\def\vc{{\bm{c}}}

\def\vf{{\bm{f}}}

\def\vp{{\bm{p}}}

\def\vt{{\bm{t}}}

\def\vx{{\bm{x}}}
\def\vy{{\bm{y}}}

\DeclareMathAlphabet{\mathsfit}{\encodingdefault}{\sfdefault}{m}{sl}
\SetMathAlphabet{\mathsfit}{bold}{\encodingdefault}{\sfdefault}{bx}{n}

\let\vec\mathbf 
\newcommand{\E}[0]{\mathbb{E}}
\newcommand{\R}[0]{\mathbb{R}}



%% file: contents/00abstract.tex
\begin{abstract}
We present Diffusion Model Patching (DMP), a simple method to boost the performance of pre-trained diffusion models that have \textit{already reached convergence}, with a negligible increase in parameters.
DMP inserts a small, learnable set of prompts into the model's input space while keeping the original model frozen.
The effectiveness of DMP is not merely due to the addition of parameters but stems from its dynamic gating mechanism, which selects and combines a subset of learnable prompts at every timestep (\ie, reverse denoising steps).
This strategy, which we term ``mixture-of-prompts'', enables the model to draw on the distinct expertise of each prompt, essentially ``patching'' the model's functionality at every timestep with minimal yet specialized parameters.
Uniquely, DMP enhances the model by further training on the original dataset already used for pre-training, even in a scenario where significant improvements are typically not expected due to model convergence.
Notably, DMP significantly enhances the FID of converged DiT-L/2 by 10.38\% on FFHQ, achieved with only a 1.43\% parameter increase and 50K additional training iterations.
\end{abstract}

%% file: contents/01introduction.tex
\section{Introduction}
\label{sec:intro}

The rapid progress in generative modeling has been largely driven by the development and advancement of diffusion models~\cite{sohl2015deep,ho2020denoising}, which have garnered considerable attention thanks to their desirable properties, such as stable training, smooth model scaling, and good mode coverage~\cite{nichol2021improved}.
Diffusion models have set new standards in generating high-quality, diverse samples that closely match the distribution of various datasets~\cite{dhariwal2021diffusion,ramesh2021zero,saharia2022image,poole2022dreamfusion}.

Diffusion models are characterized by their multi-step denoising process, which progressively refines random noise into structured outputs, such as images.
Each step aims to denoise a noised input, gradually converting completely random noise into a meaningful image.
Despite all denoising steps share the same goal of generating high-quality images, each step has distinct characteristics that contribute to shaping the final output~\cite{go2023addressing,park2023denoising}.
The visual concepts that diffusion models learn vary based on the noise ratio of input~\cite{choi2022perception}.
At higher noise levels (timestep $t$ is close to $T$), where images are highly corrupted and thus contents are unrecognizable, the models focus on recovering global structures and colors.
As the noise level decreases and images become less corrupted (timestep $t$ is close to $0$), the task of recovering images becomes more straightforward, and diffusion models learn to recover fine-grained details.
Recent studies~\cite{balaji2022ediffi,choi2022perception,go2023addressing,park2023denoising} suggest that considering \textit{stage-specificity} is beneficial, as it aligns better with the nuanced requirements of different stages in the generation process.
However, many existing diffusion models do not explicitly consider this aspect.

\input{figs/04further_training}
\input{figs/01overview}

Our goal is to enhance \textit{already converged} pre-trained diffusion models by introducing stage-specific capabilities.
We propose Diffusion Model Patching (DMP), a method that equips pre-trained diffusion models with an enhanced toolkit, enabling them to navigate the generation process with greater finesse and precision.
An overview of DMP is shown in~\cref{fig:overview}.
DMP consists of two main components:
\textbf{(1)} A small pool of learnable prompts~\cite{lester2021power}, each optimized for particular stages of the generation process.
These prompts are attached to the model's input space and act as ``experts'' for certain denoising steps (or noise levels).
This design enables the model to be directed towards specific behaviors for each stage without retraining the entire model, instead adjusting only small parameters at the input space.
\textbf{(2)} A dynamic gating mechanism that adaptively generates ``expert'' prompts (or mixture-of-prompts) based on the noise levels of the input image.
This dynamic utilization of prompts empowers the model with flexibility, enabling it to utilize distinct aspects of prompt knowledge sets at different stages of generation.
By leveraging specialized knowledge embedded in each prompt, the model can adapt to stage-specific requirements throughout the multi-step generation process.

By incorporating these components, we continue training the converged diffusion models using the original dataset on which they were pre-trained.
Given that the model has \textit{already converged}, it is generally assumed that conventional fine-tuning would not lead to significant improvements or may even cause overfitting.
However, DMP provides the model with a nuanced understanding of each denoising step, leading to enhanced performance, even when trained on the same data distribution.
As shown in~\cref{fig:further_training}, DMP boosts the performance of DiT-L/2~\cite{peebles2022scalable} by 10.38\% with only 50K iterations on the FFHQ~\cite{karras2019style} 256 $\times$ 256 dataset.


While simple, DMP offers several key strengths:

\noindent\ding{202} \textbf{Data}:
DMP boosts model performance using the original dataset, without requiring any external datasets.
This is particularly noteworthy as further training of already converged diffusion models on the same dataset typically does not lead to performance gains.
DMP differs from general fine-tuning~\cite{pan2009survey,deng2009imagenet}, which often transfers knowledge across different datasets.

\noindent\ding{203} \textbf{Computational Efficiency}:
DMP patches pre-trained diffusion models by slightly modifying their input space, without updating the model itself.
DMP contrasts with methods that train diffusion models from scratch for denoising-stage-specificity~\cite{choi2022perception,hang2023efficient,go2023addressing,park2023denoising,park2024switch}, which can be computationally expensive and storage-intensive.

\noindent\ding{204} \textbf{Parameter Efficiency}:
DMP adds only a negligible number of parameters, approximately 1.43\% of the total model parameters (based on DiT-L/2).
This ensures that performance enhancements are achieved cost-effectively.

\noindent\ding{205} \textbf{Model}:
DMP eliminates the need to train multiple expert networks for different denoising stages.
Instead, it combines a few prompts in various ways to learn nuanced behaviors specific to each step.
This simplifies the model architecture and training process compared to prior methods~\cite{balaji2022ediffi,feng2023ernie,xue2023raphael, park2024switch}.

%% file: figs/04further_training.tex
\begin{figure}[!t]
    \centering
    \includegraphics[width=0.7\linewidth]{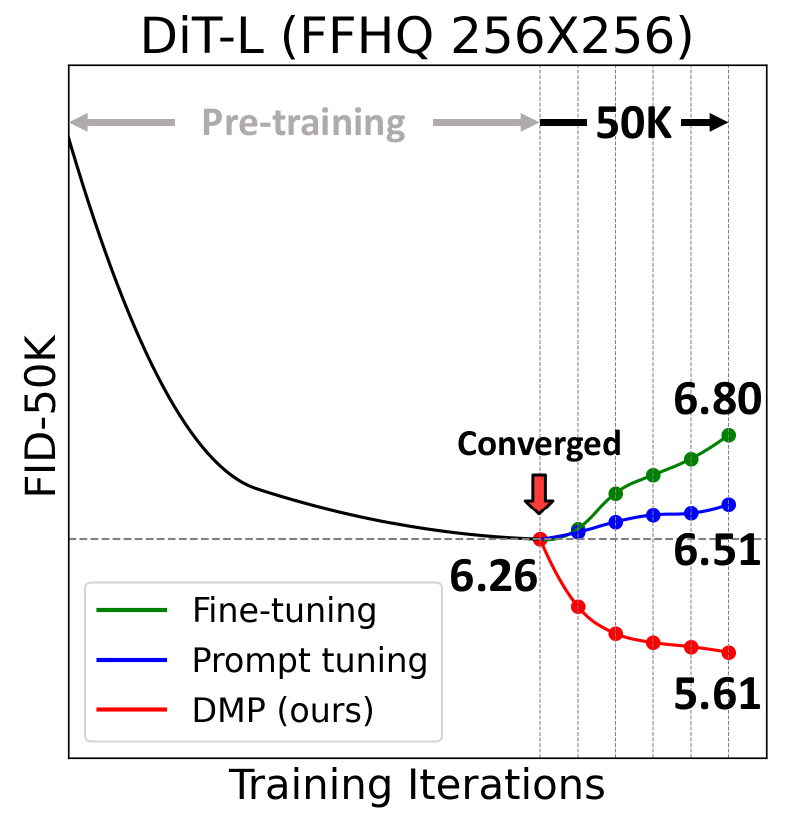}
    \vspace{-3mm}
    \captionof{figure}{
        \textbf{Further training} of the \textit{fully converged} DiT-L/2 model using the \textit{same dataset as the pre-training phase}.
        Our method, DMP achieves a 10.38\% FID improvement in just 50K iterations, while other methods exhibit overfitting.
    }
    \label{fig:further_training}
    \vspace{\belowfigcapmargin}
\end{figure}

%% file: figs/01overview.tex
\begin{figure*}[ht]
    \centering
    \includegraphics[width=0.9\linewidth]{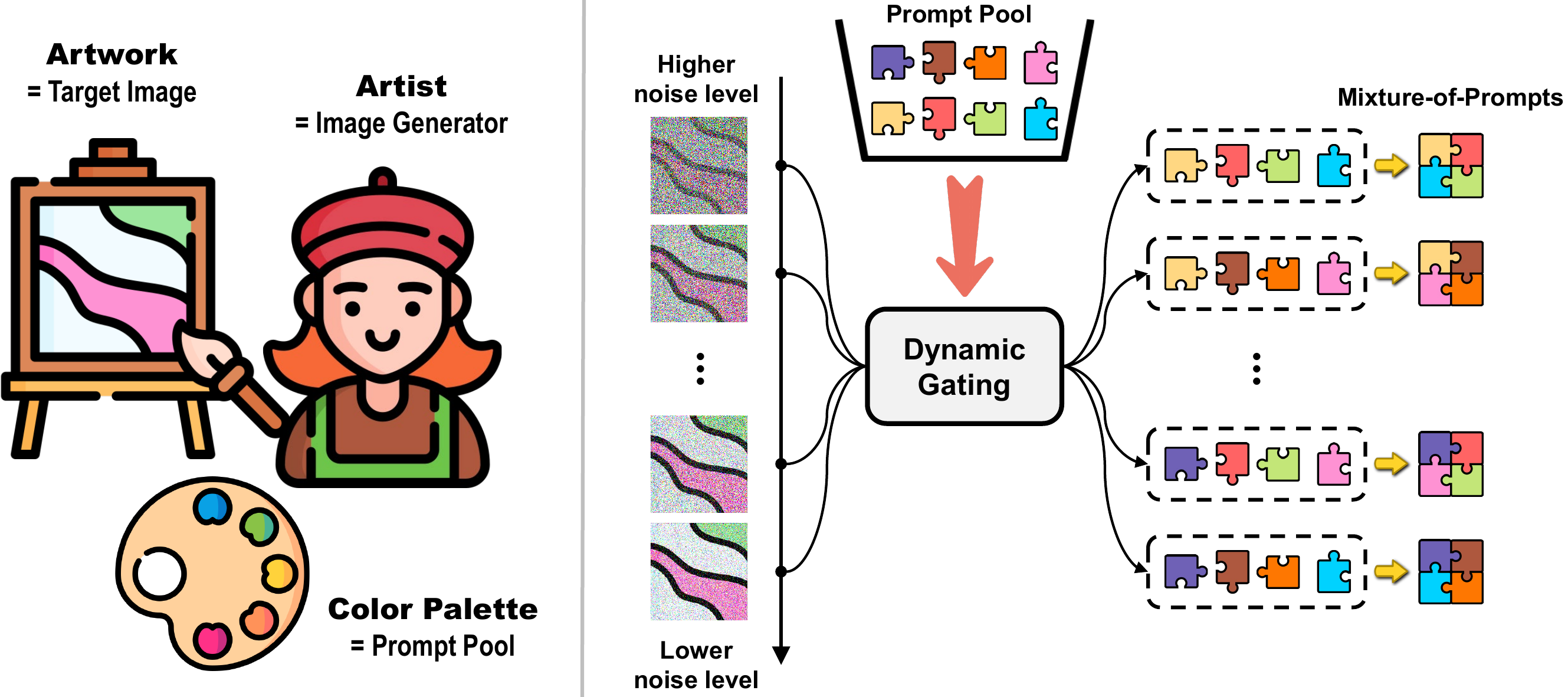}
    \vspace{-2mm}
    \caption{
    \textbf{Overview of DMP}.
    We take inspiration from prompt tuning~\cite{lester2021power} and aim to enhance \textit{already converged} diffusion models.
    Our approach incorporates a pool of prompts within the input space, with each prompt learned to excel at certain stages of the denoising process.
    At every step, a unique blend of prompts (\ie, mixture-of-prompts) is constructed via dynamic gating based on the current noise level.
    This mechanism is similar to an skilled artist choosing the appropriate color combinations to refine different aspects of their artwork for specific moments.
    Importantly, our method keeps the diffusion model itself unchanged, and only use the original training dataset for further training.
    }%
    \label{fig:overview}
    \vspace{\belowfigcapmargin}
\end{figure*}

%% file: contents/02related_work.tex
\vspace{\secmargin}\section{Related Work}
\label{sec:related_work}

\noindent\textbf{Diffusion models with stage-specificity.}
Recent advancements in diffusion models~\cite{sohl2015deep,ho2020denoising,song2020denoising} have broadened their utility across various data modalities, including images~\cite{ramesh2021zero,saharia2022photorealistic}, audios~\cite{kong2020diffwave}, texts~\cite{li2022diffusion} and 3D~\cite{woo2023harmonyview}, showcasing remarkable versatility in numerous generation tasks.
Recent efforts have focused on improving the specificity of denoising stages, with notable progress on both \textit{architectural} and \textit{optimization} fronts.
\textbf{(1)} On the \textbf{architectural} front, eDiff-I~\cite{balaji2022ediffi}, ERNIE-ViLG 2.0~\cite{feng2023ernie}, and RAPHAEL~\cite{xue2023raphael} introduced the concept of utilizing multiple expert denoisers, each tailored to specific noise levels, thereby augmenting the model's capacity.
DTR~\cite{park2023denoising} refined diffusion model architectures by allocating different channel combinations for each denoising step.
\textbf{(2)} From an \textbf{optimization} perspective, 
P2 Weight~\cite{choi2022perception} and Min-SNR Weight~\cite{hang2023efficient} accelerated convergence by framing diffusion training as a multi-task learning problem~\cite{caruana1997multitask}, where loss weights are adjusted based on task difficulty at each timestep.
Go~\etal~\cite{go2023addressing} mitigated learning conflicts of multiple denoising stages by clustering similar stages based on their signal-to-noise ratios (SNRs).
Previous studies aim to improve the specificity of denoising stages, often by assuming either training from scratch or using multiple expert networks, which can be resource-intensive and require significant parameter storage.
Whereas, our approach achieves stage-specificity \textit{without} modifying the original model parameters, starting from and using only a single pre-trained diffusion model. 

\vspace{1mm}\noindent\textbf{Parameter-efficient Fine-tuning (PEFT) in Diffusion models.}
PEFT offers a way to enhance models by tuning a small number of (extra) parameters, avoiding the need to retrain the entire model and significantly reducing computational and storage costs~\cite{xiang2023closer}.
This is particularly appealing given the complexity and parameter-dense nature of diffusion models~\cite{rombach2022high,peebles2022scalable}, where directly training diffusion models from scratch is impractical.
Recent advancements in this field can be broadly categorized into three streams:
\textbf{(1)} T2i-Adapter~\cite{mou2023t2i}, SCEdit~\cite{jiang2023scedit},  ControlNet~\cite{zhang2023adding} and CDMs~\cite{golatkar2023training} utilize \textbf{adapters}~\cite{houlsby2019parameter,hu2021lora,chen2022adaptformer} or \textbf{side-tuning}~\cite{zhang2020side,sung2022lst} to modify the neural network's behavior at specific layers.
\textbf{(2)} Textual Inversion~\cite{hertz2022prompt, gal2022image} use a concept similar to \textbf{prompt tuning}~\cite{li2021prefix,lester2021power,logan2021cutting,jia2022visual,zhou2022learning} that modifies input or textual representations to influence subsequent processing without changing the function itself.
\textbf{(3)} CustomDiffusion~\cite{kumari2023multi}, SVDiff~\cite{han2023svdiff}, and DiffFit~\cite{xie2023difffit} focus on \textbf{partial parameter tuning}~\cite{zaken2021bitfit,xu2021raise,lian2022scaling}, fine-tuning specific parameters of the neural network, such as bias terms.
These approaches have been successful in tuning diffusion models for personalization, using samples different from the original pre-training dataset. 
In contrast, our work aims to optimize the performance of pre-trained diffusion models with their original training datasets.
While being parameter-efficient, our approach targets in-domain enhancements.\footnote{Extended related work is in Appendix.}

%% file: contents/04approach.tex
\section{Diffusion Model Patching  (DMP)}
\label{sec:approach}
\input{figs/03dmp}
We propose DMP, a simple yet effective method aimed at enhancing \textit{already converged} pre-trained diffusion models by enabling them to leverage knowledge specific to different denoising stages.\footnote{See preliminaries in Appendix.}
DMP comprises two key components: (1) a pool of learnable prompts and (2) a dynamic gating mechanism.
First, a small number of learnable prompts are attached to the input space of the diffusion model.
Second, the dynamic gating mechanism selects the optimal set of prompts (or mixture-of-prompts) based on the noise levels of the input image.
Upon these components, we further train the model using the same pre-training dataset to learn prompts while keeping the backbone parameters frozen.
The overall framework of DMP is shown in~\cref{fig:dmp}.

\vspace{1mm}\noindent\textbf{Motivation.}
During the multi-step denoising process, the difficulty and purpose of each stage vary depending on the noise level~\cite{choi2022perception,hang2023efficient,go2023addressing,park2023denoising}.
Prompt tuning~\cite{li2021prefix,lester2021power,jia2022visual} assumes that if a pre-trained model already has sufficient knowledge, carefully constructed prompts can extract knowledge for a specific downstream task from the frozen model.
Likewise, we hypothesize that a pre-trained diffusion model already holds general knowledge about all denoising stages, and by learning different mixture-of-prompts for each stage, we can patch the model with stage-specific knowledge.

\vspace{1mm}\noindent\textbf{Architecture.}
As our base architecture, we employ DiT~\cite{peebles2022scalable}, which is a transformer-based~\cite{vaswani2017attention}, and Stable Diffusion~\cite{rombach2022high}, which is a UNet-based~\cite{ho2020denoising}, both operating in the latent space~\cite{rombach2022high}.
We use a pre-trained VAE~\cite{kingma2013auto} from Stable Diffusion~\cite{rombach2022high} to process input images into a latent code of shape $H \times W \times D$ (for $256\times256\times3$ images, the latent code is $32\times32\times4$). For DiT, the noisy latent code is divided into $N$ fixed-size patches, each of shape $K \times K \times D$.
Our DMP method adaptively adjusts the size of learnable prompts to the input size of each model, where $N \times D$ for DiT and $H \times W \times D$ for Stable Diffusion.

\vspace{1mm}\noindent\textbf{Learnable prompts.}
Our goal is to efficiently enhance the model with denoising-stage-specific knowledge, adjusting small parameters within the input space.
To achieve this, we start with a pre-trained DiT model as an example for explanation.  Firstly, we insert $N$ learnable continuous embeddings of dimension $D$, \ie, \textit{prompts}, into the input space of each DiT block.
The set of learnable prompts is denoted as:
\begin{align}
\vec{P} = \{\vp^{(i)} \in \R^{N \times D} \mid i \in \{0, \dots, L-1\}\}.
\label{eq:5}
\end{align}
Here, $\vp^{(i)}$ denotes the prompts for the $i$-th DiT block and $L$ is the total number of DiT blocks in the model.
Unlike previous methods~\cite{li2021prefix,lester2021power,jia2022visual,wang2022learning}, where prompts are typically prepended to the input sequence, we directly add them to the input.
This offers the advantage of not increasing the sequence length, thereby maintaining nearly the same computation speed as before.
Moreover, during the generation process, each prompt added to the input patch provides a direct signal to help denoise specific spatial parts at each timestep.
This design choice allows the model to focus on different aspects of the input at each timestep, aiding in specialized denoising steps.
The output of $i$-th DiT block is computed as follows:
\begin{align}
    \vx^{(i+1)} = {\rm Block}^{(i)}_{frozen}(\vp^{(i)}_{learn} + \vx^{(i)}).
\label{eq:6}
\end{align}
During further training, only the prompts are updated, while the DiT blocks remain unchanged. 
The subscript $learn$ indicates learnable parameters, while $frozen$ indicates frozen parameters.

\vspace{1mm}\noindent\textbf{Dynamic gating.}
In \cref{eq:6}, the same prompts are used throughout the training, thus they will learn denoising-stage-agnostic knowledge. 
To patch the model with stage-specific knowledge, we introduce dynamic gating.
This mechanism blends prompts in varying proportions based on the noise level of an input image. 
Specifically, we use a timestep embedding $\vt$ to represent the noise level at each step of the generation process.
For a given $\vt$, the gating network $\mathcal{G}$ creates the stage-specific mask of shape $N \times 1$ used for generating mixtures-of-prompts, thereby redefining \cref{eq:6} as:
\begin{align}
    \vx^{(i+1)} = {\rm Block}^{(i)}_{frozen}\big({\sigma}(\mathcal{G}_{learn}([\vt; i])) \odot \vp^{(i)}_{learn} + \vx^{(i)}\big),
\label{eq:7}
\end{align}
where $\sigma$ is the softmax function and $\odot$ denotes element-wise multiplication.
In practice, $\mathcal{G}$ is implemented as a simple linear layer.
It additionally takes the DiT block depth $i$ as input to produce different results based on the depth.
This dynamic gating mechanism effectively handles varying noise levels using only a small number of prompts.
It also provides the model with the flexibility to use different sets of prompt knowledge at different stages of the generation process.




\subsection{Training}
\noindent\textbf{Zero-initialization.}
We empirically found that random initialization of prompts disrupts the early training process, leading to instability and divergence.
To ensure stable further training of a pre-trained diffusion model, we start by zero-initializing the prompts. 
With the prompt addition strategy that we selected before, zero-initialization prevents harmful noise from affecting the deep features of neural network layers and preserves the original knowledge at the beginning of training.
As training progresses, the model gradually incorporate additional signals from the prompts.

\vspace{1mm}\noindent\textbf{Prompt balancing loss.}
We adopt two soft constraints from Shazeer~\etal~\shortcite{shazeer2017outrageously} to balance the activation of mixtures-of-prompts.
(1) Load balancing: In a mixture-of-experts setup~\cite{jacobs1991adaptive,jordan1994hierarchical}, Eigen~\etal~\cite{eigen2013learning} noted that once experts are selected, they tend to be consistently chosen.
In our setup, the load balancing loss prevents the gating network $\mathcal{G}$ from overly favoring a few prompts with larger weights and encourages all prompts to uniformly selected.
(2) Importance balancing: Despite having similar overall load, prompts might still be activated with imbalanced weights.
For instance, one prompt might be assigned with larger weights for a few denoising steps, while another might have smaller weights for many steps.
The load balancing loss ensures that prompts are activated with similar overall importance across all denoising steps.\footnote{For further details about prompt balancing loss, see Appendix.}

\input{tabs/01model_params}
\vspace{1mm}\noindent\textbf{Prompt efficiency.}
\Cref{tab:model_params} presents the model parameters for various versions of the DiT architecture~\cite{peebles2022scalable} ranging from DiT-B/2 to DiT-L/2 (where ``2'' denotes the patch size $K$) and Stable Diffusion v1.5 with and without the DMP.
Assuming a fixed 256$\times$256 resolution, using the DMP increases DiT-B/2 parameters by 1.96\%.
For the largest model, Stable Diffusion v1.5, the use of DMP results in a 0.3\% increase to 892M parameters.
The proportion of DMP parameters to total model parameters decreases as the model size increases, allowing for tuning only a small number of parameters compared to the entire model.

%% file: figs/03dmp.tex
\begin{figure}[!t]
    \centering
    \includegraphics[width=0.9\linewidth]{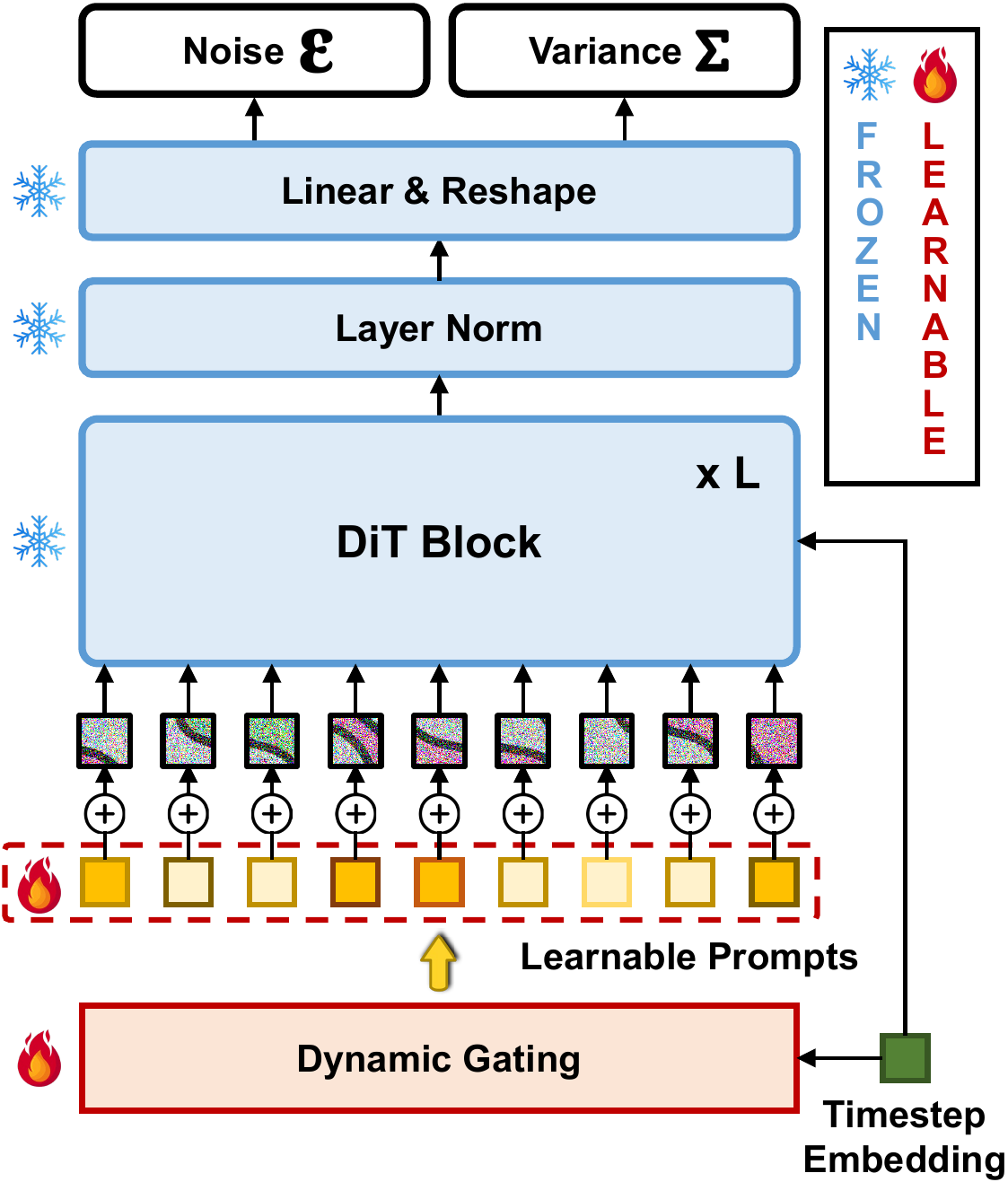}
    \vspace{-2mm}
    \caption{
        \textbf{DMP framework with DiT~\cite{peebles2022scalable}.}
        DMP is designed to adaptively generate optimal prompts tailored to specific timesteps.
        DMP uses the original training dataset---previously used for pre-training diffusion models---for fine-tuning.
        Operating entirely through prompt-based tuning in the input space, DMP eliminates the need for modifications to either the model architecture or the overall training process, ensuring seamless integration and efficiency.
    }
    \label{fig:dmp}
    \vspace{\belowfigcapmargin}
\end{figure}


%% file: tabs/01model_params.tex
\begin{table}[!t]
    \small
    \centering
                \begin{tabular}{ll}
                    \toprule
                    Model & \#Parameters \\
                    \toprule
                    DiT-B/2 & 130M \\
                    + DMP & 132.5M \scriptsize {(+1.96\%)} \\
                    \arrayrulecolor{gray}\cmidrule(lr){1-2}
                    DiT-L/2 & 458M \\
                    + DMP & 464.5M \scriptsize {(+1.43\%)} \\
                    \arrayrulecolor{gray}\cmidrule(lr){1-2}
                    SD v1.5 & 890M \\
                    + DMP & 892M \scriptsize {(+0.32\%)} \\
                    \arrayrulecolor{black}\bottomrule
                \end{tabular}
    \vspace{\abovetabcapmargin}
    \captionof{table}{
        \textbf{Parameters.} Default image size is 256$\times$256. SD v1.5 indicates Stable Diffusion v1.5~\cite{rombach2022high}.
    }
    \vspace{\belowtabcapmargin}
    \vspace{-3mm}
    \label{tab:model_params}
\end{table}

%% file: contents/05experiments.tex
\input{tabs/02main_results}
\section{Experiments}
\label{sec:experiments}
We evaluate the effectiveness of DMP on various image generation tasks using \textit{already converged} pre-trained diffusion models.
Unlike conventional fine-tuning or prompt tuning, the original dataset from the pre-training phase is used for further training.
We evaluated image quality using FID~\cite{heusel2017gans} score, which measures the distance between feature representations of generated and real images using an Inception-v3 model~\cite{szegedy2016rethinking}.
\footnote{Implementation details are in Appendix.}

\vspace{1mm}\noindent\textbf{Datasets \& Tasks.}
We used three datasets for our experiments:
(1) FFHQ~\cite{karras2019style} (for \textit{unconditional image generation}) contains 70,000 training images of human faces.
(2) MS-COCO~\cite{lin2014microsoft} (for \textit{text-to-image generation}) includes 82,783 training images and 40,504 validation images, each annotated with 5 descriptive captions.
(3) Laion5B~\cite{schuhmann2022laion} (for \textit{Stable Diffusion}) consists of 5.85B image-text pairs, which is known to be used to train Stable Diffusion~\cite{rombach2022high}.

\vspace{\subsecmargin}\subsection{Comparative Study}

\paragraph{Effectiveness of DMP.}
In~\Cref{tab:main_results}, we compare DMP against two further training baselines -- (1) full fine-tuning and (2) naive prompt tuning~\cite{lester2021power} (equivalent to \cref{eq:6}) -- across various datasets for unconditional/conditional image generation tasks.
We employ pre-trained DiT models~\cite{peebles2022scalable} that have already reached full convergence as our backbone.
To ensure that the observed enhancement is not due to cross-dataset knowledge transfer, we further train the models using the same dataset used for pre-training.
As expected, fine-tuning does not provide additional improvements to an already converged model in all datasets and even result in overfitting.
Naive prompt tuning also fails to improve performance in almost datasets and instead lead to a decrease in performance.
DMP enhances the FID across all datasets with only 30$\sim$50K iterations, enabling the model to generate images of superior quality.
This indicates that the performance gains achieved by DMP are not merely a result of increasing parameters, but rather from its novel mixture-of-prompts strategy.
This strategy effectively patches diffusion models to operate slightly differently at each denoising step.
Moreover, the significant improvement on Stable Diffusion v1.5~\cite{rombach2022high} with Laion5B~\cite{schuhmann2022laion} demonstrates DMP's expandability to different architectures and resolutions.

\input{tabs/03further_training}
\paragraph{Effects across training iterations.}
In~\Cref{tab:further_training}, further training of a DiT-L/2 model reveals interesting dynamics.
First, fine-tuning fails to increase performance beyond its already converged state and even tends to overfit, leading to performance degradation.
We also found that prompt tuning, which uses a small number of extra parameters, actually harms performance, possibly because these extra parameters act as noise in the model's input space.
In contrast, DMP, which also uses the set of parameters as prompt tuning, significantly boosts performance.
The key difference between them lies in the use of a gating function: prompts are shared across all timesteps, while DMP activates prompts differently for each timestep.
This distinction allows DMP, with a fixed number of parameters, to scale across thousands of timesteps by creating mixtures-of-prompts.
By patching stage-specificity into the pre-trained diffusion model, DMP achieves a 10.38\% FID gain in just 50K iterations.

\input{tabs/04ablations} 

\subsection{Design Choices}

\noindent\textbf{Prompt depth.}
To investigate the impact of the number of blocks in which prompts are inserted, we conduct an ablation study using the DiT-B/2 model, which comprises 12 DiT blocks.
We evaluate the performance differences when applying mixtures-of-prompts at various depths in~\cref{fig:prompt_depth}: only at the first block, up to half of the blocks, and across all blocks.
Regardless of the depth, performance consistently improves compared to the baseline with no prompts (FID=6.27).
Our findings indicate a positive correlation between prompt depth and performance, with better results achieved using mixture-of-prompts across more blocks.
The prompts selected for each block are illustrated in~\cref{fig:prompt_activation}.


\vspace{1mm}\noindent\textbf{Gating architecture.}
Our DMP framework incorporates a dynamic gating mechanism to select mixture-of-prompts.
We compare the impact of two gating architectures in Table \ref{tab:gating_architecture}: linear gating \vs attention gating.
The linear gating utilizes a single linear layer, taking a timestep embedding as input to produce a weighting mask for each learnable prompt.
On the other hand, attention gating utilizes an attention layer~\cite{vaswani2017attention}, treating learnable prompts as a query and timestep embeddings as key and value,
resulting in weighted prompts directly.
Upon comparing the two gating architectures, we found that linear gating achieves better FID (5.87) compared to attention gating (6.41).
As a result, we adopt linear gating as our default setting.

\input{figs/05prompt_activation}
\input{figs/07qualitative}
\vspace{1mm}\noindent\textbf{Gating type.}
In Table \ref{tab:gating_type}, we evaluate two design choices for creating mixture-of-prompts: hard  \vs soft selection.
With hard selection, we choose the top-$192$ prompts out of 256 prompts (75\% of total prompts) based on the gating probabilities, using them only with a weight of 1.
Whereas, soft selection uses all prompts but assigns different weights to each.
Soft selection leads to further improvement, whereas hard selection results in worse performance.
Therefore, we set the soft selection as our default setting.

\vspace{1mm}\noindent\textbf{Prompt selection.}
By default, DMP inserts learnable prompts into the input space of every blocks in the diffusion model.
Two choices arise in this context:
(1) uniform: gating function $\mathcal{G}$ in \cref{eq:7} receives only the timestep embedding $\vt$ as input and applies common weights to the prompts at every block, thus prompt selection is consistent across all blocks.
(2) distinct: $\mathcal{G}$ processes not only $\vt$ but also current block depth $i$ as inputs, generating different weights for each block.
As shown in Table \ref{tab:prompt_selection}, using distinct prompt selections leads to enhanced performance.
Therefore, we input both the timestep embedding and current block depth information to the gating function, enabling the distinct prompt combinations for each block depth as our default setting.

\vspace{1mm}\noindent\textbf{Prompt position.}
While previous prompt tuning approaches~\cite{li2021prefix,lester2021power,jia2022visual,zhou2022learning,wang2022learning} typically prepend learnable prompts to image tokens, we directly add prompts element-wise to image tokens to maintain the input sequence length. 
Table \ref{tab:prompt_position} ablates different choices for inserting prompts into the input space and their impact on performance.
We compare two methods: prepend \vs add.
For ``add'', we use 256 prompts to match the number of image tokens, and for ``prepend'', we utilize 50 prompts.
Although "prepend" should ideally use 256 tokens for a fair comparison, we limit it to 50 tokens due to severe divergence, even when both methods are equally zero-initialized.
The results show that ``add'' method improves performance with a stable optimization process, achieving a 5.87 FID compared to the baseline of 6.27 FID.
On the other hand, the ``prepend'' leads to a drop in performance, with an FID of 6.79. 
Additionally, ``add'' has the advantage of not increasing computation overhead.
Based on these findings, we set ``add'' as our default for stable optimization.

\vspace{1mm}\noindent\textbf{Prompt balancing.}
Prompt balancing loss acts as a soft constraint for the gating function, mitigating biased selections of prompts when producing a mixture-of-prompts.
We study the impact of two types of balancing losses by altering the coefficient values for load balancing loss and importance balancing loss.
As shown in Table \ref{tab:prompt_balancing}, using both types of losses equally enables the diffusion model to reach its peak performance.
This indicates that balancing both the number and weight of the activated prompts across different timesteps is crucial for creating an effective mixture-of-prompts. 
Consequently, we employ equal proportions of importance balancing and load balancing losses for prompt balancing.




\subsection{Analysis}


\noindent\textbf{Prompt activation.}
The gating function plays a pivotal role in dynamically crafting mixtures-of-prompts from a set of learnable prompts, based on the noise level present in the input.
This is depicted in \cref{fig:prompt_activation}, where the activation is visually highlighted using colors.
As the denoising process progresses, the selection of prompts exhibits significant variation across different timesteps.
At higher timesteps with high noise levels, the gating function tends to utilize a broader array of prompts.
Conversely, at lower timesteps, as the noise diminishes, the prompts become more specialized, focusing narrowly on specific features of the input that demand closer attention.
This strategic deployment of prompts allows the model to form specialized "experts" at each denoising step, catering to the specific needs dictated by the input's noise characteristics and enhancing the model's performance.

\vspace{1mm}\noindent\textbf{Qualitative analysis.}
\cref{fig:qualitative} presents a visual comparison between three methods: the baseline DiT model, prompt tuning~\cite{lester2021power}, and our DMP.
These methods are evaluated on unconditional, text-to-image generation tasks using FFHQ~\cite{karras2019style} and COCO~\cite{lin2014microsoft}, respectively.
DMP generates realistic and natural images with fewer artifacts. 


\vspace{1mm}\noindent\textbf{Additional results and analysis.} Appendix provides a theoretical grounding of DMP, experiments on training diffusion models from scratch with DMP, applying DMP on DiT-XL/2, comparisons with LoRA~\cite{hu2021lora}, and ablations of gating conditions. It also examines the structural bias of DMP and provides additional qualitative results.

%% file: tabs/02main_results.tex
\begin{table*}[t]
    \begin{center}
        \begin{large}
        \setlength\tabcolsep{6pt}
            \scalebox{0.81}{
                \begin{tabular}{lc}
                \toprule
                Resolution ($256\times256$) & FFHQ \\
                \arrayrulecolor{gray}\cmidrule(lr){2-2}
                Model & FID{\down}\\
                \arrayrulecolor{black}\midrule 
                \multicolumn{2}{l}{\textit{Pre-trained (iter: 600K)}} \\ 
                \quad DiT-B/2 & 6.27 \\
                \arrayrulecolor{gray}\cmidrule(lr){1-2}
                \multicolumn{2}{l}{\textit{Further Training (iter: 30K)}} \\ 
                \quad + Fine-tuning   & 6.57$_{{(+0.30)}}$ \\
                \quad + Prompt tuning & 6.81$_{{(+0.54)}}$ \\
                \arrayrulecolor{gray}\cmidrule(lr){1-2}
                \quad \textbf{+ DMP}& \cellcolor{Gray}\textbf{5.87$_{\textbf{(-0.40)}}$} \\
                \arrayrulecolor{black}\bottomrule
                \end{tabular}
            }
            \scalebox{0.81}{
                \begin{tabular}{lc}
                \toprule
                Resolution ($256\times256$) & COCO \\
                \arrayrulecolor{gray}\cmidrule(lr){2-2}
                Model & FID{\down} \\
                \arrayrulecolor{black}\midrule 
                \multicolumn{2}{l}{\textit{Pre-trained (iter: 450K)}} \\ 
                \quad DiT-B/2 & 7.33 \\
                \arrayrulecolor{gray}\cmidrule(lr){1-2}
                \multicolumn{2}{l}{\textit{Further Training (iter: 40K)}} \\ 
                \quad + Fine-tuning   & 7.51$_{{(+0.18)}}$ \\
                \quad + Prompt tuning & 7.37$_{{(+0.04)}}$ \\
                \arrayrulecolor{gray}\cmidrule(lr){1-2}
                \quad \textbf{+ DMP}& \cellcolor{Gray}\textbf{7.12$_{\textbf{(-0.21)}}$} \\
                \arrayrulecolor{black}\bottomrule
                \end{tabular}
            }
            \scalebox{0.81}{
                \begin{tabular}{lc}
                \toprule
                Resolution ($512\times512$) & {Laion5B} \\
                \arrayrulecolor{gray}\cmidrule(lr){2-2}
                Model & FID{\down} \\
                \arrayrulecolor{black}\midrule 
                \multicolumn{2}{l}{\textit{Pre-trained \textit{(iter: 1.5M)}}} \\ 
                \quad Stable Diffusion v1.5 & 47.18 \\
                \arrayrulecolor{gray}\cmidrule(lr){1-2}
                \multicolumn{2}{l}{\textit{Further Training (iter: 50K)}} \\ 
                \quad + Fine-tuning   & 52.99$_{{(+5.81)}}$ \\
                \quad + Prompt tuning & 35.93$_{{(-11.25)}}$ \\
                \arrayrulecolor{gray}\cmidrule(lr){1-2}
                \quad \textbf{+ DMP} & \cellcolor{Gray}\textbf{35.44$_{\textbf{(-11.74)}}$} \\
                \arrayrulecolor{black}\bottomrule
                \end{tabular}
            }
            \vspace{-2mm}
            \captionof{table}{
                \textbf{Evaluating pre-trained diffusion models with different further training methods.}
                Importantly, we use the same dataset as in the pre-training for further training.
                We set two baselines for comparison:
                (1) \textit{full fine-tuning} to update the entire model parameters.
                (2) \textit{naive prompt tuning}~\cite{lester2021power} (equivalent to~\cref{eq:6}).
                {\down}: The lower the better.
            }
            \label{tab:main_results}
            \vspace{\belowtabcapmargin}
        \end{large}
    \end{center}
    \vspace{-5pt}
\end{table*}

%% file: tabs/03further_training.tex
\begin{table}[!t]
    \begin{center}
        \begin{small}
            \setlength\tabcolsep{2pt}
            \scalebox{0.95}{
                \begin{tabular}{lcccccl}
                    \toprule
                    FFHQ $256\times256$ & \multicolumn{6}{c}{Further training iterations} \\ 
                    \arrayrulecolor{gray}\cmidrule(lr){1-7}
                    {DiT-L/2 \textit{(iter: 250K)}} & {+0K} & {+10K} & {+20K} & {+30K} & {+40K} & {+50K} \\
                    \arrayrulecolor{gray}\cmidrule(lr){2-7}
                    \quad + Fine-tuning & {6.26} & {6.32} & {6.53} & {6.64} & {6.73} & 6.80$_{(+0.54)}$ \\
                    \quad + Prompt tuning & {6.26} & {6.30} & {6.36} & {6.40} & {6.42} & 6.51$_{(+0.25)}$ \\
                    \arrayrulecolor{gray}\cmidrule(lr){1-7}
                    \quad \textbf{+ DMP} & {6.26} & {5.88} & {5.72} & {5.67} & {5.64} & \cellcolor{Gray}\textbf{5.61$_{(-0.65)}$} \\
                    \arrayrulecolor{black}\bottomrule
                \end{tabular}
            }
            \vspace{\abovetabcapmargin}
            \caption{\textbf{Comparison of further training techniques across iterations} on FFHQ 256$\times$256.
            }
            \vspace{-4mm}
            \vspace{\belowtabcapmargin}
            \label{tab:further_training}
        \end{small}
    \end{center}
\end{table}

%% file: tabs/04ablations.tex
\begin{figure*}[t!]
    \begin{minipage}[t!]{0.27\linewidth}
        \begin{center}
        \vspace{-2mm}
        \includegraphics[width=\linewidth]{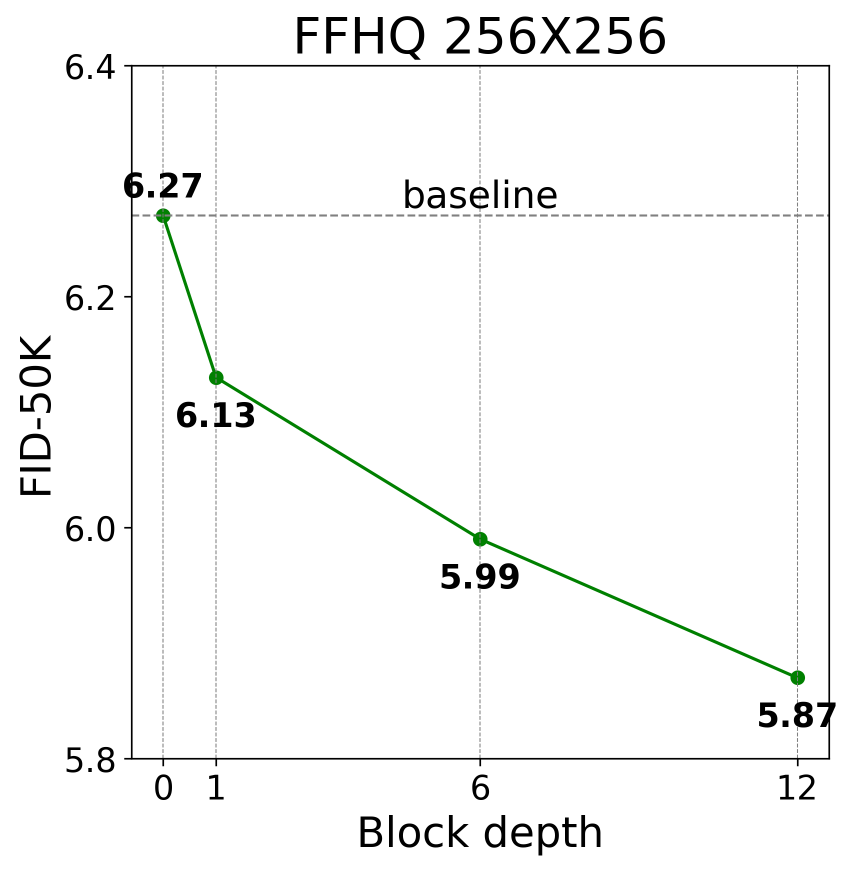}
        \end{center}
        \vspace{-5mm}
        \captionof{figure}{
            \textbf{Prompt depth.}
        }
        \label{fig:prompt_depth}
    \end{minipage}
    \hfill
    \begin{minipage}[t!]{0.72\linewidth}
        \begin{center}
            \begin{small}
                \begin{minipage}{0.66\linewidth}{
                    \begin{minipage}{0.48\linewidth}{\begin{center}
                    \tablestyle{10pt}{1.05}
                    \begin{subtable}{\linewidth}
                        \centering
                        \begin{tabular}{l|c}
                            {case} & FID{\down} \\
                            \shline
                            attention & 6.41 \\
                            \textbf{linear} & \cellcolor{Gray}\textbf{5.87} \\
                        \end{tabular}
                        \caption{\scriptsize\textbf{Gating architecture.}}
                        \label{tab:gating_architecture}
                    \end{subtable}
                    \end{center}}\end{minipage}
                    \hfill
                    \begin{minipage}{0.48\linewidth}{\begin{center}
                    \tablestyle{10pt}{1.05}
                    \begin{subtable}{\linewidth}
                        \centering
                        \begin{tabular}{l|c}
                            {case} & FID{\down} \\
                            \shline
                            hard & 5.96 \\
                            \textbf{soft} & \cellcolor{Gray}\textbf{5.87} \\
                        \end{tabular}
                        \caption{\scriptsize\textbf{Gating type.}}
                        \label{tab:gating_type}
                    \end{subtable}
                    \end{center}}\end{minipage}
                    \hfill
                    \begin{minipage}{0.48\linewidth}{\begin{center}
                    \tablestyle{10pt}{1.05}
                    \begin{subtable}{\linewidth}
                        \centering
                        \begin{tabular}{l|c}
                            {case} & FID{\down} \\
                            \shline
                            uniform & 5.97 \\
                            \textbf{distinct} & \cellcolor{Gray}\textbf{5.87} \\
                        \end{tabular}
                        \caption{\scriptsize\textbf{Prompt selection.}}
                        \label{tab:prompt_selection}
                    \end{subtable}
                    \end{center}}\end{minipage}
                    \hfill
                    \begin{minipage}{0.48\linewidth}{\begin{center}
                    \tablestyle{10pt}{1.05}
                    \begin{subtable}{\linewidth}
                        \centering
                        \begin{tabular}{l|c}
                            {case} & FID{\down} \\
                            \shline
                            prepend & 6.79 \\
                            \textbf{add} & \cellcolor{Gray}\textbf{5.87} \\
                        \end{tabular}
                        \caption{\scriptsize\textbf{Prompt position.}}
                        \label{tab:prompt_position}
                    \end{subtable}
                    \end{center}}\end{minipage}
                }\end{minipage}
                \begin{minipage}{0.32\linewidth}{
                    \begin{minipage}{\linewidth}{\begin{center}
                    \tablestyle{10pt}{1.05}
                    \begin{subtable}{\linewidth}
                        \centering
                        \scalebox{0.99}{
                        \begin{tabular}{cc|c}
                            {IB} & {LB} & FID{\down} \\
                            \shline
                            0 & 0 & 6.11\\ 
                            0 & 1 & 5.96\\ 
                            1 & 0 & 5.97\\ 
                            1 & 2 & 5.95\\
                            2 & 1 & 5.95\\ 
                            \textbf{1} & \textbf{1} & \cellcolor{Gray}\textbf{5.87}\\
                        \end{tabular}}
                        \caption{\scriptsize\textbf{Prompt balancing.} IB: importance balancing; LB: load balancing.}
                        \label{tab:prompt_balancing}
                    \end{subtable}
                    \end{center}}\end{minipage}
                }\end{minipage}
            \captionof{table}{
                \textbf{DMP ablations.}
                DiT-B/2~\cite{peebles2022scalable} pre-trained on FFHQ 256$\times$256~\cite{karras2019style} is further trained for 30K iterations with DMP (Baseline FID = 6.27). {\down}: The lower the better.
            }
            \label{tab:ablation}
            \vspace{\belowtabcapmargin}
            \end{small}
        \end{center}
    \end{minipage}
    \vspace{-10pt}
\end{figure*}

%% file: figs/05prompt_activation.tex
\begin{figure}[t!]
        \centering
        \includegraphics[width=\linewidth]{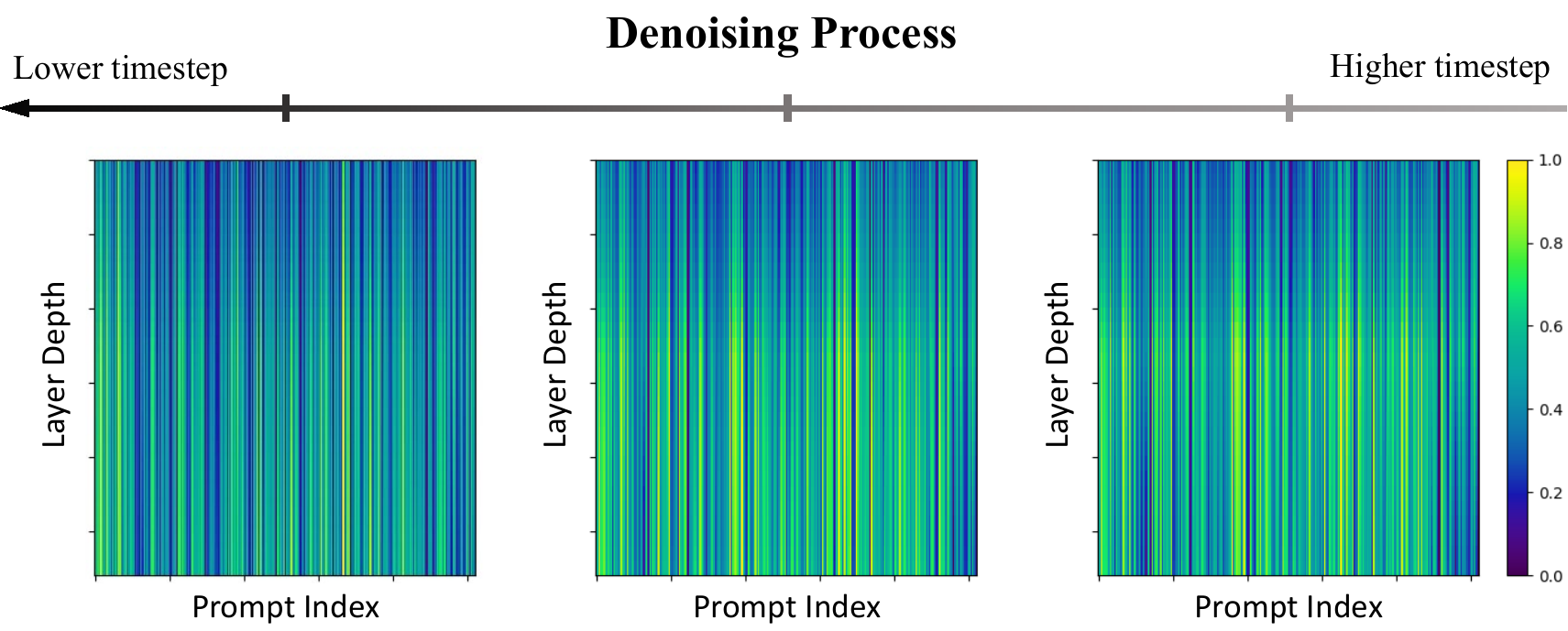}
        \vspace{\abovefigcapmargin}
        \captionof{figure}{
            \textbf{Prompt activation.}
            Brighter indicates stronger.
        }
        \label{fig:prompt_activation}
        \vspace{\belowfigcapmargin}
\end{figure}

%% file: figs/07qualitative.tex
\begin{figure*}[t!]
    \centering
        \setlength{\tabcolsep}{5pt}
        \renewcommand{\arraystretch}{0.5}
        \begin{tabular}{z{1}x{55}x{55}x{55}x{55}x{55}x{55}x{55}} 
        \multicolumn{8}{l}{\footnotesize\textbf{(a) Unconditional Image Generation on FFHQ~\cite{karras2019style}}} \\
        \arrayrulecolor{white}\midrule 
        
        \raisebox{0.5\height}{\rotatebox{90}{\footnotesize Baseline}} &
        \adjincludegraphics[clip,width=0.125\textwidth,trim={0 0 0 0}]{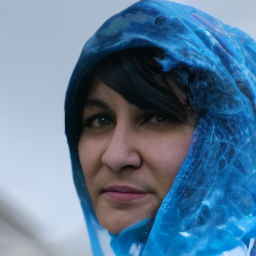} &
        \adjincludegraphics[clip,width=0.125\textwidth,trim={0 0 0 0}]{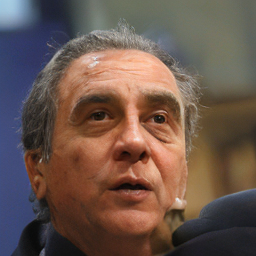} &
        \adjincludegraphics[clip,width=0.125\textwidth,trim={0 0 0 0}]{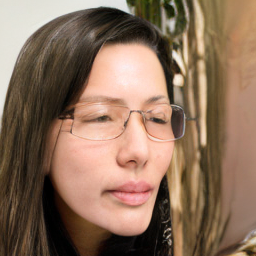} &
        \adjincludegraphics[clip,width=0.125\textwidth,trim={0 0 0 0}]{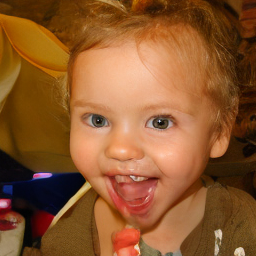} &
        \adjincludegraphics[clip,width=0.125\textwidth,trim={0 0 0 0}]{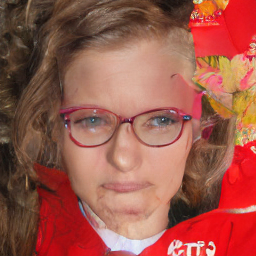} &
        \adjincludegraphics[clip,width=0.125\textwidth,trim={0 0 0 0}]{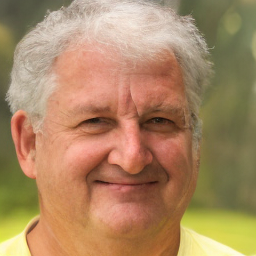} &
        \adjincludegraphics[clip,width=0.125\textwidth,trim={0 0 0 0}]{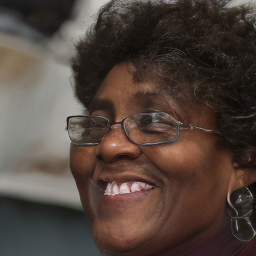} \\
        
        \raisebox{1.2\height}{\rotatebox{90}{\footnotesize + PT}} &
        \adjincludegraphics[clip,width=0.125\textwidth,trim={0 0 0 0}]{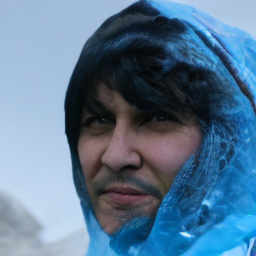} &
        \adjincludegraphics[clip,width=0.125\textwidth,trim={0 0 0 0}]{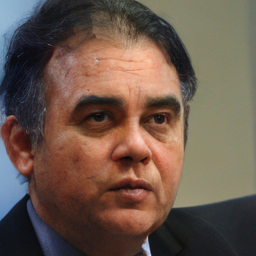} &
        \adjincludegraphics[clip,width=0.125\textwidth,trim={0 0 0 0}]{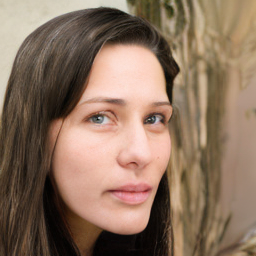} &
        \adjincludegraphics[clip,width=0.125\textwidth,trim={0 0 0 0}]{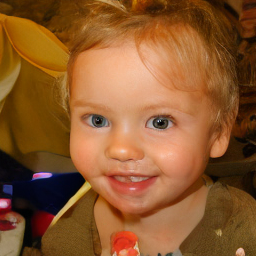} &
        \adjincludegraphics[clip,width=0.125\textwidth,trim={0 0 0 0}]{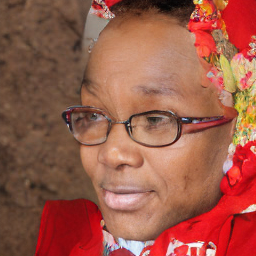} &
        \adjincludegraphics[clip,width=0.125\textwidth,trim={0 0 0 0}]{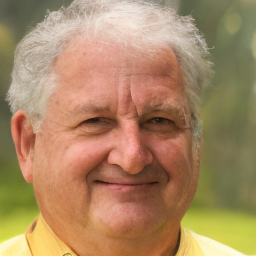} &
        \adjincludegraphics[clip,width=0.125\textwidth,trim={0 0 0 0}]{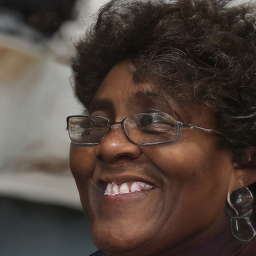} \\
        
        \raisebox{0.7\height}{\rotatebox{90}{\footnotesize + DMP}} &
        \adjincludegraphics[clip,width=0.125\textwidth,trim={0 0 0 0}]{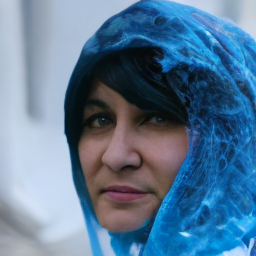} &
        \adjincludegraphics[clip,width=0.125\textwidth,trim={0 0 0 0}]{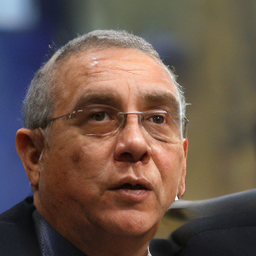} &
        \adjincludegraphics[clip,width=0.125\textwidth,trim={0 0 0 0}]{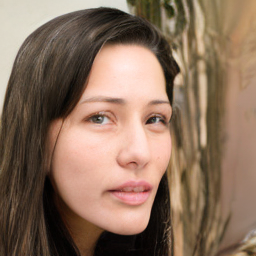} &
        \adjincludegraphics[clip,width=0.125\textwidth,trim={0 0 0 0}]{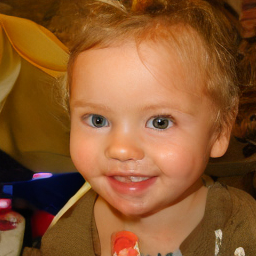} &
        \adjincludegraphics[clip,width=0.125\textwidth,trim={0 0 0 0}]{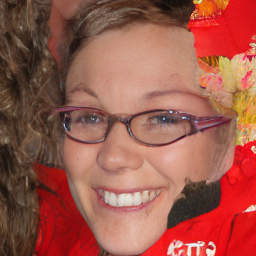} &
        \adjincludegraphics[clip,width=0.125\textwidth,trim={0 0 0 0}]{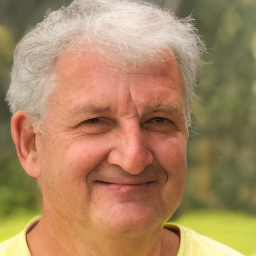} &
        \adjincludegraphics[clip,width=0.125\textwidth,trim={0 0 0 0}]{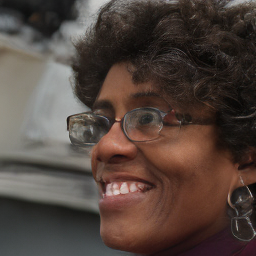} \\

        \arrayrulecolor{white}\midrule
        \multicolumn{8}{l}{\footnotesize\textbf{(b) Text-to-Image Generation on MS-COCO~\cite{lin2014microsoft}}.} \\
        \arrayrulecolor{white}\midrule 
        \raisebox{0.5\height}{\rotatebox{90}{\footnotesize Baseline}} &
        \adjincludegraphics[clip,width=0.125\textwidth,trim={0 0 0 0}]{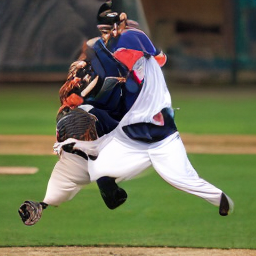} &
        \adjincludegraphics[clip,width=0.125\textwidth,trim={0 0 0 0}]{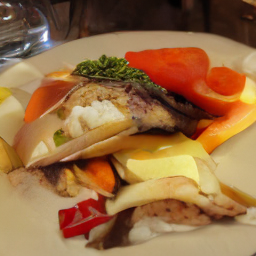} &
        \adjincludegraphics[clip,width=0.125\textwidth,trim={0 0 0 0}]{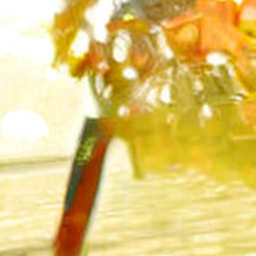} &
        \adjincludegraphics[clip,width=0.125\textwidth,trim={0 0 0 0}]{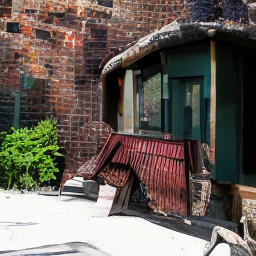} &
        \adjincludegraphics[clip,width=0.125\textwidth,trim={0 0 0 0}]{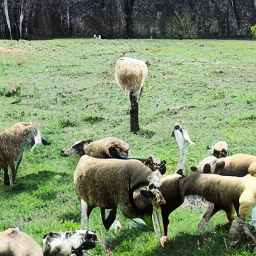} &
        \adjincludegraphics[clip,width=0.125\textwidth,trim={0 0 0 0}]{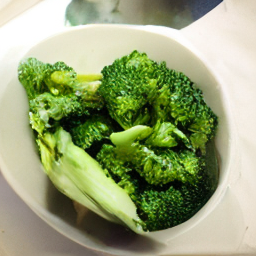} &
        \adjincludegraphics[clip,width=0.125\textwidth,trim={0 0 0 0}]{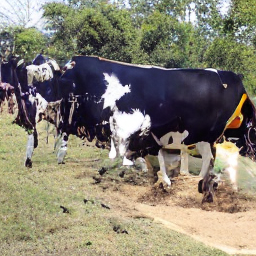} \\
        
        \raisebox{1.2\height}{\rotatebox{90}{\footnotesize + PT}} &
        \adjincludegraphics[clip,width=0.125\textwidth,trim={0 0 0 0}]{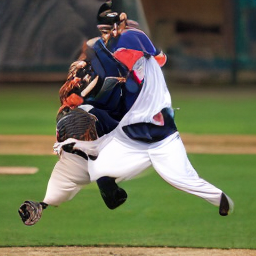} &
        \adjincludegraphics[clip,width=0.125\textwidth,trim={0 0 0 0}]{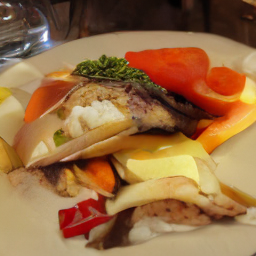} &
        \adjincludegraphics[clip,width=0.125\textwidth,trim={0 0 0 0}]{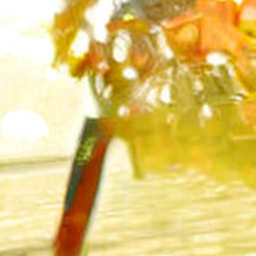} &
        \adjincludegraphics[clip,width=0.125\textwidth,trim={0 0 0 0}]{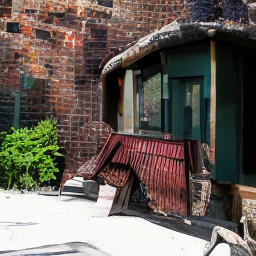} &
        \adjincludegraphics[clip,width=0.125\textwidth,trim={0 0 0 0}]{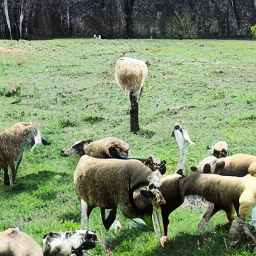} &
        \adjincludegraphics[clip,width=0.125\textwidth,trim={0 0 0 0}]{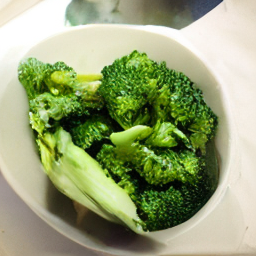} &
        \adjincludegraphics[clip,width=0.125\textwidth,trim={0 0 0 0}]{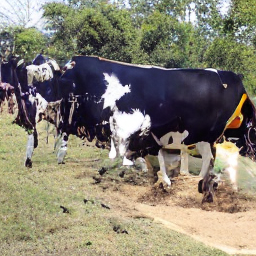} \\
        
        \raisebox{0.7\height}{\rotatebox{90}{\footnotesize + DMP}} &
        \adjincludegraphics[clip,width=0.125\textwidth,trim={0 0 0 0}]{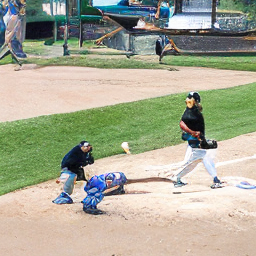} &
        \adjincludegraphics[clip,width=0.125\textwidth,trim={0 0 0 0}]{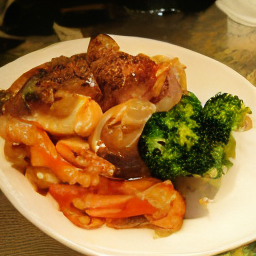} &
        \adjincludegraphics[clip,width=0.125\textwidth,trim={0 0 0 0}]{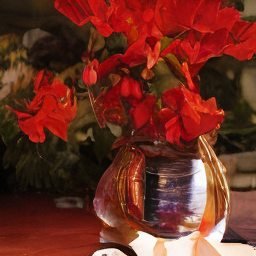} &
        \adjincludegraphics[clip,width=0.125\textwidth,trim={0 0 0 0}]{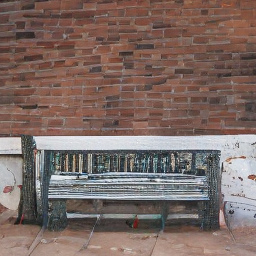} &
        \adjincludegraphics[clip,width=0.125\textwidth,trim={0 0 0 0}]{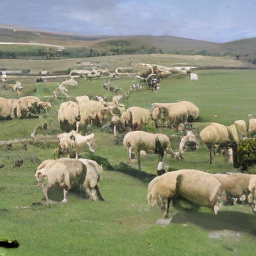} &
        \adjincludegraphics[clip,width=0.125\textwidth,trim={0 0 0 0}]{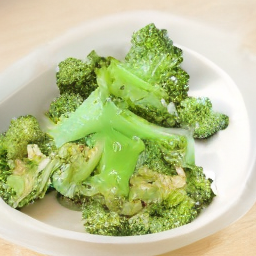} &
        \adjincludegraphics[clip,width=0.125\textwidth,trim={0 0 0 0}]{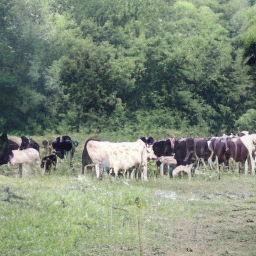} \\
        & {\begin{spacing}{1}\tiny A man getting ready to catch a baseball.\end{spacing}} & {\begin{spacing}{1}\tiny A white plate with vegetables underneath sliced up meat.\end{spacing}} & {\begin{spacing}{1}\tiny Red vase with yellow flowers sticking out of it.\end{spacing}} & {\begin{spacing}{1}\tiny A bench sitting in front of a brick wall on a patio.\end{spacing}} & {\begin{spacing}{1}\tiny A herd of sheep gathered in one area.\end{spacing}} & {\begin{spacing}{1}\tiny Cut green broccoli florets in a white serving bowl.\end{spacing}} & {\begin{spacing}{1}\tiny Black and white cows stand around in a farm yard.\end{spacing}}
    \end{tabular}
    \vspace{-5mm}
    \vspace{\abovetabcapmargin}
    \caption{
        \textbf{Qualitative comparison} among the baseline (DiT-B/2~\cite{peebles2022scalable}), naive prompt tuning (PT) applied to the baseline, and DMP applied to the baseline on (a) FFHQ and (b) MS-COCO datasets.
    }
    \vspace{\belowtabcapmargin}
    \label{fig:qualitative}
\end{figure*}


%% file: contents/06conclusion.tex
\section{Conclusion}
\label{sec:conclusion}
We introduced Diffusion Model Patching (DMP), a simple method for further enhancing pre-trained diffusion models that have already converged. 
By incorporating timestep-specific learnable prompts and leveraging dynamic gating, DMP adapts the model’s behavior dynamically across thousands of denoising steps.
This design enables DMP to effectively address the variations inherent in denoising stages, which are often overlooked in existing diffusion model architectures.
Our results demonstrate that DMP achieves significant performance gains without the need for extensive retraining.
Applied to the DiT-L/2 backbone, DMP delivered a 10.38\% improvement in FID after just 50K iterations, with a minimal parameter increase of 1.43\% on the FFHQ 256×256 dataset.
Additionally, its adaptability across different models and image generation tasks underscores its potential as a versatile enhancement method for diffusion models.

%% file: contents/08appendix.tex
\section*{\centering\LARGE Appendix}
\appendix

\section{Theoretical Grounding of DMP}
Our work is grounded in the framework of Multi-task Learning~\cite{caruana1997multitask}, specifically leveraging parameter sharing and parameter separation notions. We detail the theoretical framework for our approach:

\noindent 1) \textbf{Parameter Sharing}: We build upon an already converged diffusion model, parameterized by $\theta$, which serves as the shared trunk. This shared trunk has been trained across all timesteps $t \in T$, effectively functioning as a multi-task model by learning a general representation applicable to all tasks (timesteps).

\noindent 2) \textbf{Parameter Separation}: To enhance performance, we introduce new prompts parameterized by $\phi$. These prompts are designed to learn specialized parameters for each timestep, thereby explicitly addressing the unique aspects of each task within the multi-task learning framework.

Mathematically, let $x_t$ denote the input at timestep $t$, and let $f(x_t; \theta)$ represent the output of the shared trunk. The introduction of prompts modifies this output, which can be expressed as:
\begin{equation}
    y_t = f(x_t; \theta) + g(x_t, t; \phi)
\end{equation}
where $g(x_t, t; \phi)$ represents the contribution of the prompt-specific parameters.
During training, the shared trunk parameters $\theta$ remain fixed, while the prompt parameters $\phi$ are optimized. This ensures that the learning process focuses on fine-tuning the model for each specific timestep without altering the foundational multi-task model. The gate mechanism dynamically adjusts the weight of each prompt’s contribution based on the timestep, effectively performing parameter separation. This can be represented as:
\begin{equation}
    \alpha_t = \text{Gate}(t)
\end{equation}
\begin{equation}
    y_t = f(x_t; \theta) + \alpha_t \cdot g(x_t, t; \phi)
\end{equation}
where $\alpha_t$ is the weight determined by the gate for timestep $t$.

By reusing the original data used for pre-training (in-domain) to learn the prompts, we ensure that the performance improvements are due to the Multi-task learning effect rather than transfer learning.

\section{Extended Related Work}
\label{sec:additional_related_work}
\vspace{\paramargin}\paragraph{Prompt-based learning.}
Recent progress in NLP has shifted towards leveraging pre-trained language models (LMs) using textual prompts~\cite{petroni2019language,radford2019language,schick2020s,liu2023pre} to guide models to perform target tasks or produce desired outputs without additional task-specific training.
With strategically designed prompts, models like GPT-3~\cite{brown2020language} have shown impressive generalization across various downstream tasks, even under few-shot or zero-shot conditions.
Prompt tuning~\cite{li2021prefix,lester2021power,liu2023gpt,liu2021p} treats prompts as learnable parameters optimized with supervision signals from downstream training samples while keeping the LM's parameters fixed.
Similar principles have also been explored in visual~\cite{jia2022visual,bahng2022exploring} and vision-and-language~\cite{zhou2022learning,zhou2022conditional} domains.
To the best of our knowledge, there is currently no direct extension of prompt tuning in enhancing the in-domain performance of diffusion models.
While Prompt2prompt~\cite{hertz2022prompt} and Textual Inversion~\cite{gal2022image} share similar properties with prompt tuning, their focus is on customized editing or personalized content generation.
In this work, we propose to leverage prompt tuning to enhance the stage-specific capabilities of diffusion models.
With only a small number of prompts, we can effectively scale to thousands of denoising steps via a mixture-of-prompts strategy.

\vspace{\secmargin}\section{Preliminaries}
\label{sec:preliminary}

\vspace{4pt}\vspace{\paramargin}\paragraph{Diffusion models.}
Diffusion models~\cite{dhariwal2021diffusion,song2020denoising} generate data by reversing a pre-defined diffusion process (or \textit{forward process}), which sequentially corrupts the original data $\vx_0$ into noise over a series of steps $t \in \{1, \dots, T\}$.
\begin{equation}
q(\vx_{t}|\vx_{t-1}) = \mathcal{N}(\vx_t; \sqrt{1-\beta_t}\vx_{t-1}, \beta_t\mathbf{I}),
\label{eq:1}
\end{equation}
where $0 < \beta_t < 1$ is a variance schedule controlling the amount of noise added at each step.
This process results in data that resembles pure noise \(\mathcal{N}(0, \mathbf{I})\) at step $T$ (often $T=1000$).
The \textit{reverse process} aims to reconstruct the original data by denoising, starting from noise and moving backward to the initial state $\vx_0$.
This is modeled by a neural network parameterized by $\bm{\theta}$ that learns the conditional distribution $p_{\bm{\theta}}(\vx_{t-1}|\vx_t)$.
The network is trained by optimizing a weighted sum~\cite{ho2020denoising} of denoising score matching losses~\cite{vincent2011connection} over multiple noise scales~\cite{song2019generative}.
In practice, the network predicts the noise $\bm{\epsilon}$ added at each forward step, rather than directly predicting $\vx_{t-1}$ from $\vx_t$, using the objective function:
\begin{equation}
\mathcal{L}_{t} := \E_{{\vx}_0, {\bm{\epsilon} \sim \mathcal{N}(0, 1)}, t \sim U[1, T]} \| {\bm{\epsilon}} - {\bm{\epsilon}}_{\bm{\theta}}({\vx}_t, t) \|_2^2.
\label{eq:2}
\end{equation}
By minimizing~\cref{eq:2} for all $t$, the neural network learns to effectively reverse the noising process, thereby enabling itself to generate samples from $p_{\bm{\theta}}(\vx_0)$ that closely resemble the original data distribution.

\vspace{\paramargin}\paragraph{Prompt tuning.}
The core idea behind prompt tuning is to find a small set of parameters that, when combined with the input, effectively ``tune'' the output of a pre-trained model towards desired outcomes.
Traditional fine-tuning aims to minimize the gap between ground truth $\vy$ and prediction $\hat{\vy}$ by \textit{modifying the pre-trained model ${\vf}_{\bm\theta}$}, given the input $\vx$:
\begin{equation}
\hat{\vy}' = {{\vf}^{\textit{learn}}_{\bm\theta'}}({\vx}),
\label{eq:3}
\end{equation}
where $\hat{\vy}'$ is the refined prediction, ${\vf}_{\bm\theta'}$ is the modified model, and The superscript $learn$ indicates learnable parameters, while $frozen$ indicates frozen parameters.
This process is often computationally expensive and resource-intensive, as it requires storing and updating the full model parameters.
In contrast, prompt tuning aims to enhance the output $\hat{\vy}$ by directly \textit{modifying the input $\vx$}:
\begin{equation}
\hat{\vy}' = {{\vf}^{\textit{frozen}}_{\bm\theta}}({\vx}^{learn}_{p}).
\label{eq:4}
\end{equation}
Previous works~\cite{li2021prefix,lester2021power,jia2022visual,zhou2022learning,wang2022learning} commonly define ${{\vx}^{learn}_p} = [{{\vp}^{learn}}; {{\vx}}]$, where $[\cdot; \cdot]$ denotes concatenation.
However, we take a different approach by directly adding prompts to the input, aiming to more explicitly influence the input itself, thus ${{\vx}^{learn}_{p}} = {{\vp}^{learn}} + {{\vx}}$.
Prompts are optimized via gradient descent, similar to conventional fine-tuning, but without changing the model's parameters.

\section{Architecture of Diffusion Models}
\vspace{4pt}\vspace{\paramargin}\paragraph{Architecture of DiT.}
The DiT model~\cite{peebles2022scalable} is a diffusion model that utilizes a transformer-based~\cite{vaswani2017attention} DDPM~\cite{ho2020denoising}, operating within the latent space for image generation tasks. The architecture begins by employing a pre-trained Variational Autoencoder (VAE)~\cite{kingma2013auto} from Stable Diffusion~\cite{rombach2022high} to encode input images into latent codes of shape $H \times W \times D$. For example, an image with dimensions $256\times256\times3$ is encoded into a latent code of size $32\times32\times4$.

Noise is then added to the latent code, which is a standard training procedure in diffusion models, allowing the model to learn the process of denoising. The noisy latent code is divided into $N$ fixed-size patches, each of which is linearly embedded with shape $K \times K \times D$. Positional encodings~\cite{vaswani2017attention} are added to these embedded patches, transforming them into a sequence of vectors that serve as the input tokens for the transformer model.

The core of the DiT architecture consists of a series of $L$ DiT blocks. Each block includes Multi-Head Self-Attention~\cite{vaswani2017attention}, Feed-Forward Networks, Layer Normalization~\cite{ba2016layer}, and residual connections~\cite{he2016deep}. These components work together to allow the model to focus on different parts of the input sequence, capture complex dependencies, and stabilize the training process. The blocks are conditioned with timestep embeddings $\vt$ , which provide the model with information about the stage of the denoising process, and can also be optionally conditioned with class or text embeddings.
After the last DiT block, the noisy latent patches undergo the final Layer Normalization and are linearly decoded into a $K \times K \times 2D$ tensor ($D$ for noise prediction and another $D$ for diagonal covariance prediction).
Finally, the decoded tokens are rearranged to match the original shape $H \times W \times D$.

\vspace{4pt}\vspace{\paramargin}\paragraph{Architecture of Stable Diffusion.}
Stable Diffusion~\cite{rombach2022high} employs a UNet-based architecture~\cite{ho2020denoising} for its diffusion model, also operating in the latent space to achieve efficient and high-quality image generation. Similar to the DiT model, Stable Diffusion starts by using a pre-trained VAE to encode input images into latent codes. For instance, an image of size $256\times256\times3$ is transformed into a latent code of size $32\times32\times4$.

Noise is added to the latent codes according to a specific timestep sampled from the PNDM timestep scheduler~\cite{liu2022pseudo}, which is crucial for the model to learn the denoising process during training. The noisy latent codes are then processed through a series of attention blocks~\cite{vaswani2017attention} within the UNet architecture. These blocks include Multi-Head Self-Attention, which allows the model to focus on important features within the latent representation, and Cross-Attention, which is particularly important in Stable Diffusion for integrating conditioning information like text or class embeddings into the image generation process. Feed-Forward Networks, Layer Normalization, and residual connections are also used within these blocks to enhance learning capacity, stabilize the training process, and maintain gradient flow.

After processing through the attention blocks, the latent representation undergoes a final Layer Normalization and is then decoded back to its original shape through a convolutional layer, effectively reconstructing the latent code with the original shape $H \times W \times D$.

\begin{figure*}[t!]
    \centering
    \includegraphics[width=0.8\linewidth]{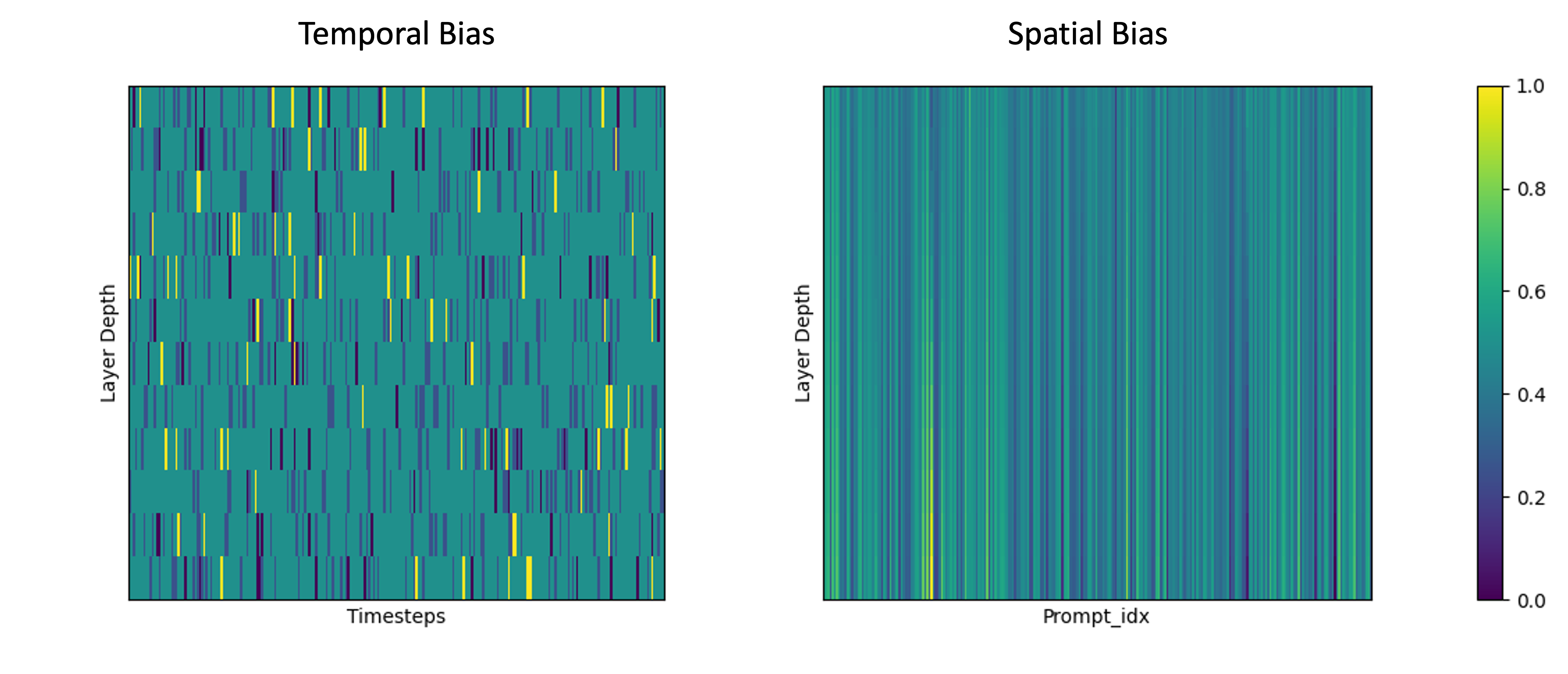}
    \vspace{-14pt}
    \caption{\textbf{Structural bias of prompts.} Temporal bias is calculated by a patch-wise average of prompt activations over 1000 timesteps. Spatial bias is calculated by a timestep-wise average of prompt activations over 256 patches.}
    \vspace{-4pt}
    \label{fig:fig1}
\end{figure*}

\section{Prompt Balancing Loss}
\label{sec:prompt_balancing_loss}
During the training stage of our DMP method, we adopt a prompt balancing loss to prevent the gate from selecting only a few specific prompts, a problem known as mode collapse. The prompt balancing loss is inspired by the balancing loss used in mixture-of-experts~\cite{shazeer2017outrageously}.
Our prompt balancing loss includes the load loss, which prevents the existence of unselected prompts, and the importance loss, which ensures the selected prompts have uniform weights.
When defining the $n$-th prompt gating weights in the $i$-th DiT-block layer as $g^i_n$, the formulations of the load balancing loss and importance loss are as follows:
\begin{equation}
    L_{Load} = \frac{1}{L}\sum^{L-1}_{i=0}\sum^{N-1}_{n=0}{\mathbb{I}(g^i_n<0)},
\end{equation}
\begin{equation}
\begin{split}
  &\Tilde{g}_n = \sum^{L-1}_{i=0}{g^i_n}, \\
  \mu = \frac{1}{N}\sum^{N-1}_{n=0}{\Tilde{g}_n}, &\quad
  \sigma = \frac{1}{N}\sum^{N-1}_{n=0}{(\Tilde{g}_n-\mu)^2},
\end{split}
\end{equation}
\begin{equation}
    L_{importance} = {\frac{\sigma}{\mu^2+\epsilon}},
\end{equation}
where $L$ is the total number of DiT block layers, $N$ is the total number of prompts, $\epsilon$ is $1e^{-5}$ to prevent division by zero, and $\mathbb{I}$ is the indicator function that counts the number of elements satisfying the condition $(g^i_n<0)$. The importance loss uses the squared coefficient of variation, which makes the variance value robust against the mean value. With these prompt balancing losses, we regularize the prompt selection in the gate, ensuring that there are few unselected prompts, as shown in Fig. \ref{fig:prompt_activation} of main manuscript.

\section{Implementation Details}
\label{sec:implementation_details}


Our experiments followed this setup for further training of pretrained DiT-B/L models~\cite{peebles2022scalable} and Stable Diffusion v1.5 model~\cite{rombach2022high}. Firstly, we used a diffusion timestep T of 1000 for training and DDPM with 250 steps~\cite{ho2020denoising} for DiT models and PNDM~\cite{liu2022pseudo} for Stable Diffusion v1.5 model during sampling. For beta scheduling, cosine scheduling~\cite{nichol2021improved} was used for the FFHQ~\cite{karras2019style} and MS-COCO datasets~\cite{lin2014microsoft}, while linear scheduling was used for the ImageNet dataset~\cite{deng2009imagenet} and Laion5B~\cite{schuhmann2022laion}. For text-to-image generation (MS-COCO) and class-conditional image generation (ImageNet) tasks, we adopted classifier-free guidance~\cite{ho2022classifier} with a guidance scale of 1.5, while for Stable Diffusion v1.5 (Laion5B), we used a guidance scale of 7.5. The batch size was set to 128, and random horizontal flipping was applied to the training data. We used the AdamW optimizer~\cite{loshchilov2017decoupled} with a fixed learning rate of 1e-4 and no weight decay.
Originally, the exponential moving average (EMA) technique is utilized for stable training of DiT models. However, since our method involves further training on pretrained models with only a few training steps, we did not adopt the EMA strategy. All experiments were conducted using a single NVIDIA A100 GPU.

\section{Structural Bias of DMP}
Our DMP approach, which adds identical prompts to image tokens at corresponding positions, has the potential to introduce a structural bias. To evaluate the extent and impact of this bias, we conducted the following analyses on DiT-B/2 model~\cite{peebles2022scalable} with FFHQ dataset~\cite{karras2019style}:
We averaged the attention weights across the same patch positions over different images throughout the entire temporal axis (\textit{i.e.}, across different timesteps). This helps us understand if the same patches in different images receive consistent attention, indicating a temporal bias.
Similarly, averaging across the same timestep positions over different images allows us to evaluate if different patches within the same timestep receive consistent attention, revealing a spatial bias.

\noindent 1) \textbf{Temporal bias}: Our analysis shows that the same patches across different images at various timesteps do exhibit some level of consistent attention. This indicates that there is a temporal structural bias introduced by the prompts.

\noindent 2) \textbf{Spatial bias}: Similarly, within the same timestep, different patches across various images also exhibit consistent attention patterns, indicating a spatial structural bias.

Note that the strength of the structural bias varies depending on the specific configurations and the dataset used. 

\paragraph{Visualization.}
In~\cref{fig:fig1}, we show the prompt activation (activated by gating mechanism conditioned on the timestep) averaged across multiple samples, which might indicate a type of bias in the ``input”. In our setup, all parameters are fixed except for the prompt, which is added directly to the image patch token. However, this might not fully reflect the impact on the “final image”.
Fundamentally, the role of the prompts in DMP is not to learn completely new information. During fine-tuning, we train on the same data distribution as in pre-training. Therefore, the prompts activated at each timestep are meant to learn the fine-grained, timestep-specific features that the pre-trained diffusion model might have missed. Hence, it does not introduce entirely new structural patterns unseen in a well-converged diffusion model.
Empirically, in our qualitative observations, we did not find any noticeable correlations or significant structural biases in the image introduced by specific prompts compared to the larger structural bias inherent in the multi-task diffusion model's shared trunk. No regular patterns or meaningful structural biases were observed in the final images.

\paragraph{Statistics.}
To numerically study the impacts of prompts on image-level, we conducted an extended analysis focusing on the spatial dimensions of the latent representations.
The statistics are shown in~\Cref{tab:tab1}.
We generated 1,000 latent representations for both the baseline DiT-B/2 model and the DiT-B/2 + DMP. Each latent representation had a shape of (4, 32, 32), where 4 represents the number of channels, and 32x32 corresponds to the spatial dimensions. We computed the Pearson correlation coefficient matrix for each set of 1,000 latent representations, resulting in a (1000, 1000) correlation matrix. This matrix captures the pairwise correlation between all latent representations within the set. To focus on unique pairs, we excluded the diagonal elements (which represent self-correlation) and computed the max, min, mean, and standard deviation of the off-diagonal elements.
Similarly, we normalized each latent representation vector and computed the cosine similarity matrix, also of shape (1000, 1000). Excluding the diagonal, we calculated the max, min, mean, and standard deviation for the off-diagonal elements.
Notably, the mean correlation and cosine similarity values are lower in the DiT-B model with DMP compared to the baseline. The reduction in mean correlation and cosine similarity implies that the latent representations generated by the DiT-B + DMP are less correlated and less similar to each other compared to those from the baseline. This suggests that DMP may contribute to enhanced diversity in the generated samples.

\input{tabs/appendix/01}

\input{tabs/appendix/00}

\section{Additional Experiments}
\paragraph{Training diffusion models with DMP from scratch.}
In~\Cref{tab:from_scratch}, we show how DMP performs when applied from the start of training.
The results indicate that applying DMP from early training stages is still effective, potentially benefiting multi-task learning (MTL) by promoting more nuanced task-specific adaptation via dynamic gating and mixture-of-prompts.

\paragraph{Comparsion with LoRA.}
We compare our method with LoRA~\cite{hu2021lora} in \Cref{tab:tab4}.
The results show that training a pre-trained diffusion model with LoRA on the original dataset used for pre-training can negatively impact the image generation capability of the diffusion model.
In contrast to our DMP, LoRA cannot inherently adjust parameters differently for each timestep. Our DMP, however, enables dynamic control of parameters at each timestep, allowing for more granular adaptation than what is achievable with traditional adapters. 
This demonstrates that our prompt-based method offers distinct advantages, particularly in terms of ease of implementation and timestep-specific adaptation.

\input{tabs/appendix/02}

\paragraph{Applying DMP to a large DiT model.}
\Cref{tab:ditxl} shows the results on the DiT-XL/2 model~\cite{peebles2022scalable}, pre-trained with the ImageNet dataset~\cite{deng2009imagenet}.
When using the same dataset from the pre-training phase for further training, both fine-tuning and prompt-tuning methods fail to enhance, and even degrade, the class-conditional image generation performance. In contrast, our DMP method effectively improves the model’s performance.
In addition, DMP maintains strong precision and recall, matching the base DiT-XL/2 model, while achieving better FID.
The improvements were achieved with only 1.26\% of additional parameters and a significantly shorter training duration (20K iterations) compared to the full training schedule (7M iterations). This efficiency highlights the practicality and cost-effectiveness of our method, especially for large models where training resources are substantial.

\input{tabs/appendix/03}

\paragraph{Ablation study on gating condition.}
In class-conditional image generation and text-to-image generation tasks, conditional guidance plays a crucial role in determining the outcome of generated images~\cite{dhariwal2021diffusion,ho2022classifier}.
To study the impact of conditional guidance on selecting mixtures-of-prompts, we evaluate the performances of two cases: one where the gating function $\mathcal{G}$ receives only the timestep embedding $\vt$, and the other where it receives both $\vt$ and the class or text condition embedding $\vc$.
In the latter case, we modify the input of the gating function in Eq. \ref{eq:7}, from $\vt$ to $\vt + \vc$.
Our analysis, presented in~\cref{tab:gating_condition}, indicates that on the ImageNet dataset, both methods perform equivalently, whereas on the COCO dataset, using $\vt + \vc$ yields superior performance compared to using $\vt$ alone.
This suggests that incorporating conditional guidance can help in determining how to combine prompts at each denoising step to generate an image that aligns well with the text condition.
Consequently, we use $\vt + \vc$ as the default input for the gating function in the text-to-image task.

\input{tabs/appendix/04}

\begin{figure*}[p]
    \includegraphics[width=\textwidth]{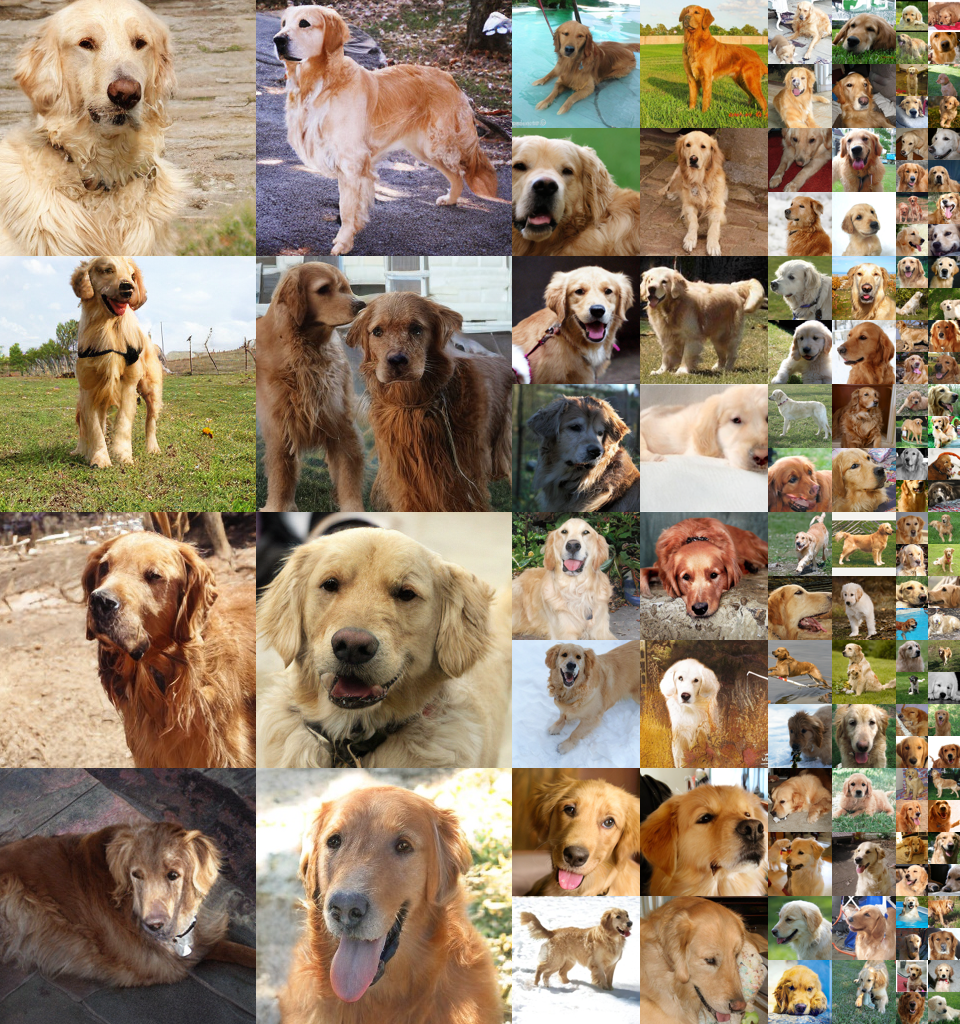}
    \captionsetup{justification=centering}
    \caption{
        \textbf{Uncurated 256×256 DiT-XL/2+DMP samples.} \\
        Classifier-free guidance scale = 1.5. \\
        Class label = “golden retriever” (207)
    }
    \label{fig:golden_retriever}
\end{figure*}

\begin{figure*}[p]
    \includegraphics[width=\textwidth]{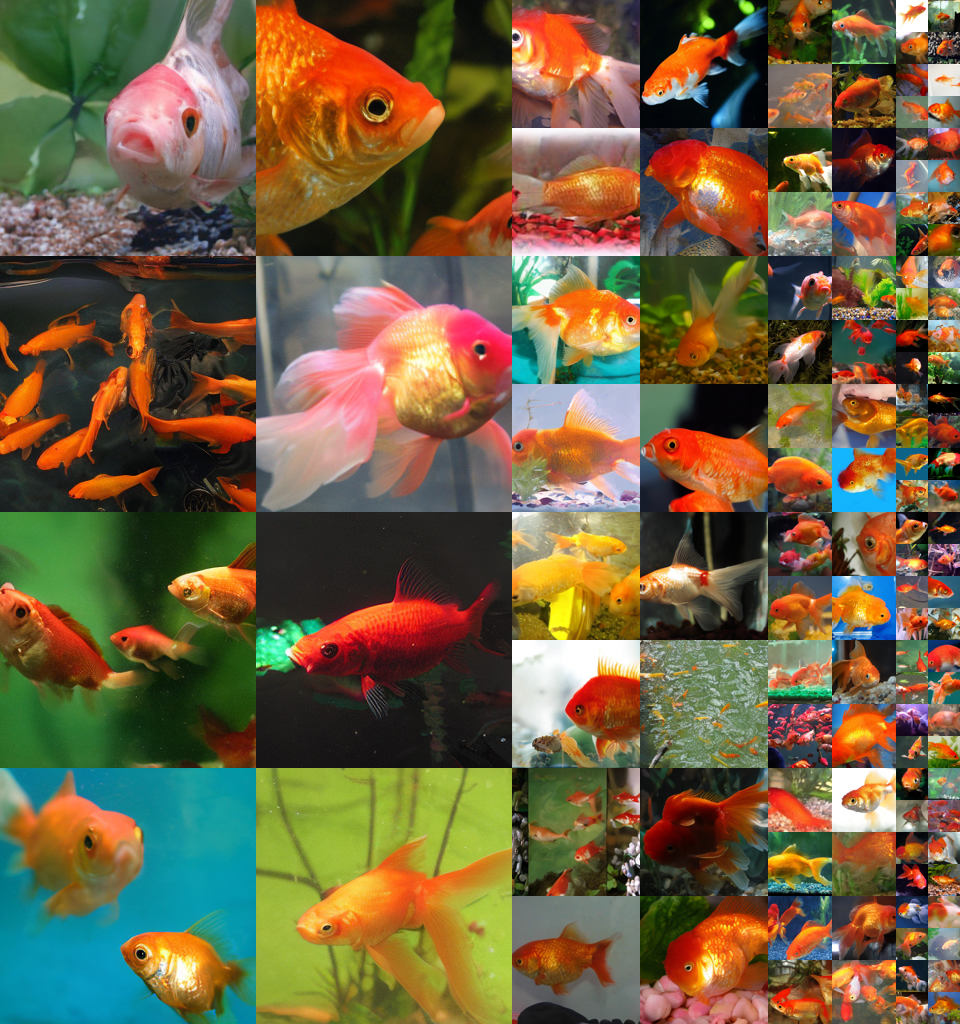}
    \captionsetup{justification=centering}
    \caption{
        \textbf{Uncurated 256×256 DiT-XL/2+DMP samples.} \\
        Classifier-free guidance scale = 1.5. \\
        Class label = “goldfish” (1)
    }
    \label{fig:goldfish}
\end{figure*}

\begin{figure*}[p]
\centering
    \includegraphics[width=\textwidth]{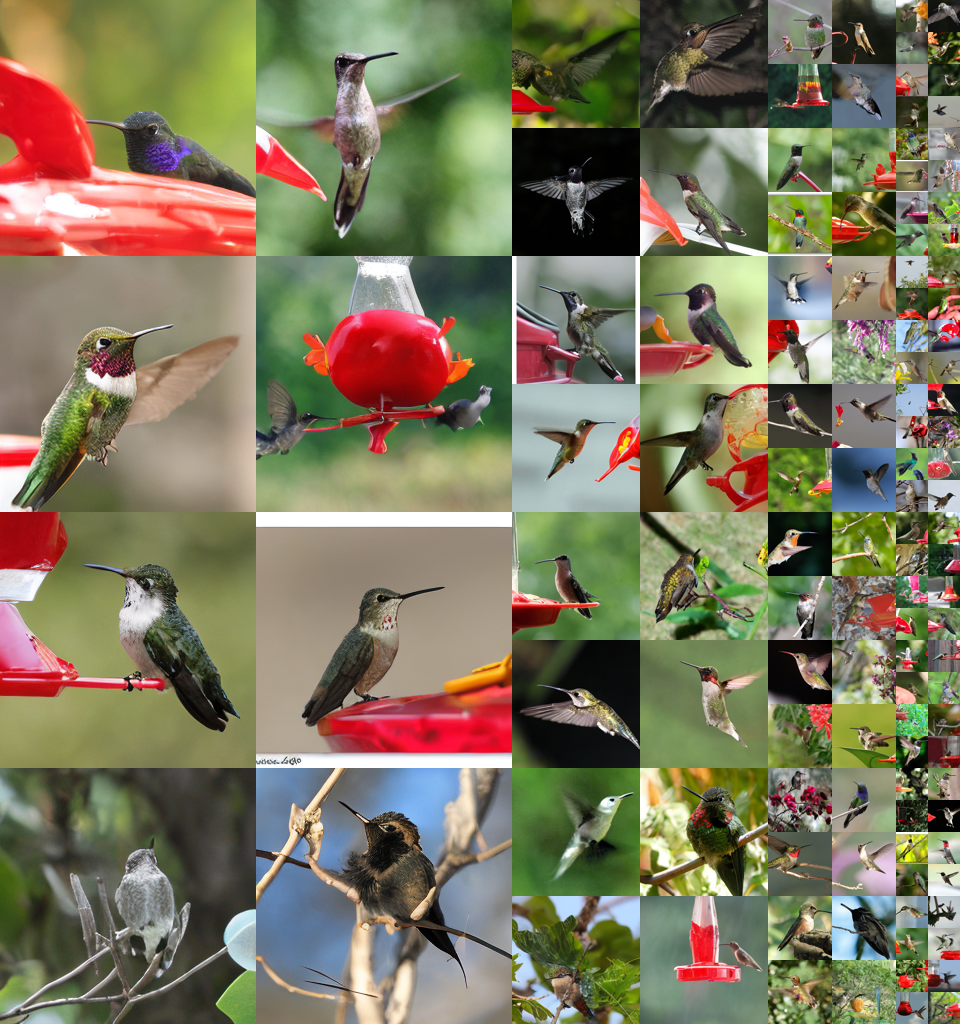}
    \captionsetup{justification=centering}
    \caption{
        \textbf{Uncurated 256×256 DiT-XL/2+DMP samples.} \\
        Classifier-free guidance scale = 1.5. \\
        Class label = “hummingbird” (94)
    }
    \label{fig:hummingbird}
\end{figure*}

\begin{figure*}[p]
\centering
    \includegraphics[width=\textwidth]{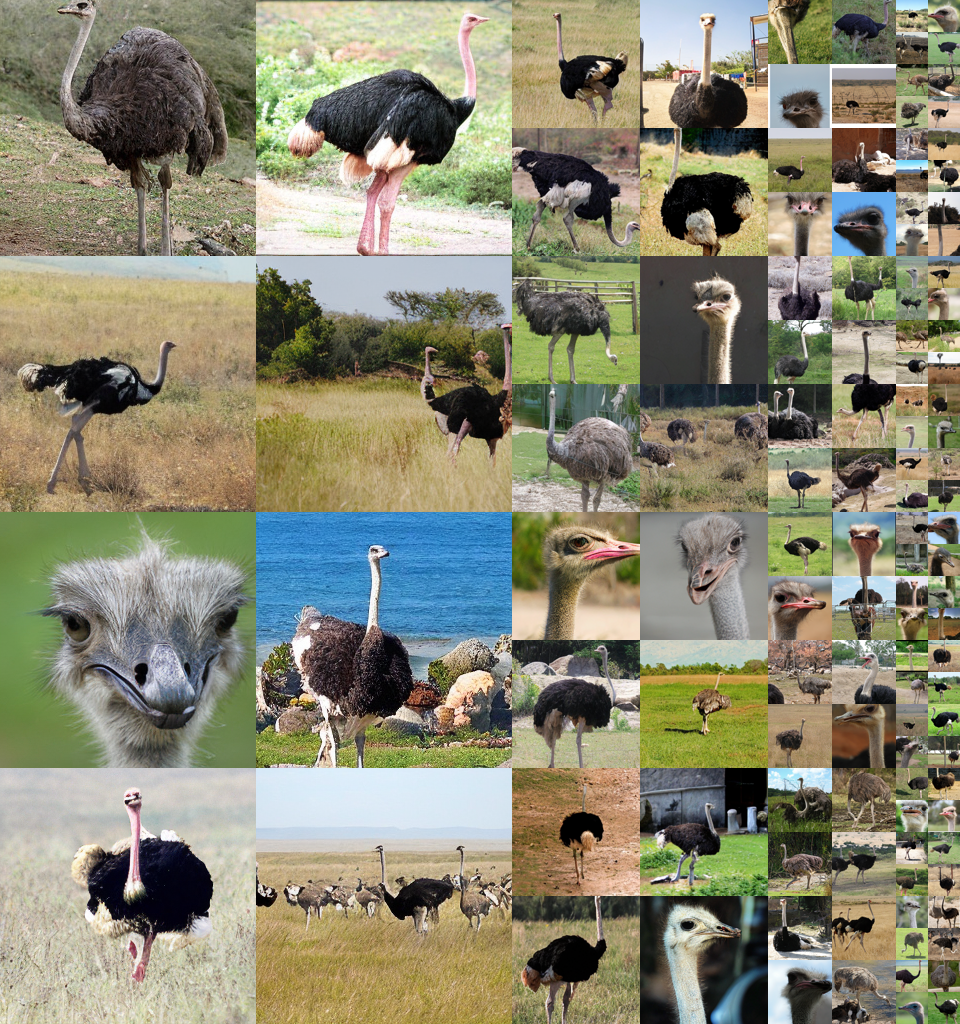}
    \captionsetup{justification=centering}
    \caption{
        \textbf{Uncurated 256×256 DiT-XL/2+DMP samples.} \\
        Classifier-free guidance scale = 1.5. \\
        Class label = “ostrich” (9)
    }
    \label{fig:ostrich}
\end{figure*}

\paragraph{Additional qualitative results on ImageNet.}
\label{sec:additional_results}
\Cref{fig:golden_retriever,fig:goldfish,fig:hummingbird,fig:ostrich} illustrates the generated images by pre-trained DiT-XL with DMP trained on 20K iterations.
The results demonstrate that highly realistic images can be generated, even with relatively limited training.

\section{Limitations}

\paragraph{Fixed number of prompts.}
Our DMP method adopts a prompt-adding strategy to ensure stable training and maintain sampling speed. However, since the number of input patches is fixed, the flexibility in the number of prompts is limited. Extending our DMP with a prepend approach while maintaining stable training is an interesting future direction.

\paragraph{Different resolutions and aspect ratio.}
Our current method, DMP, has limitations in handling different image resolutions and aspect ratios, especially in comparison to models like the FiT~\cite{lu2024fit}, Lumina-T2X~\cite{gao2024lumina}, or Any-size-diffusion~\cite{zheng2024any}, which aim to generalize across arbitrary resolutions and aspect ratios.
Unlike these prior works, our DMP requires retraining with different sized prompts specifically tailored to the target resolution and aspect ratio to achieve effective adaptation. While it is possible to use pooling or interpolation techniques to reshape the prompt size for different resolutions and aspect ratios using a pretrained prompt, this approach is arbitrary and can lead to limited performance. Therefore, developing a more flexible method for adapting DMP to varying resolutions and aspect ratios remains an area as further exploration and improvement.

%% file: tabs/appendix/01.tex
\begin{table}[t!]
\centering
\setlength{\tabcolsep}{4pt} 
\scalebox{1}{
\begin{tabular}{lc|ccccc}
Model                      & Metric      & Max    & Min     & Mean   & Std    \\ \shline
\multirow{2}{*}{DiT-B/2}     & Correlation & 0.7064 & -0.4424 & 0.1715 & 0.1172 \\
                           & Similarity  & 0.7443 & -0.4676 & 0.1764 & 0.1234 \\
                           \hline 
\multirow{2}{*}{+DMP} & Correlation & 0.7285 & -0.4770 & 0.1410 & 0.1202 \\ 
                           & Similarity  & 0.7410 & -0.5084 & 0.1366 & 0.1240 \\ 
\end{tabular}
}
\vspace{-3mm}
\caption{\textbf{The statistics of Structural bias.}}
\vspace{-3mm}
\label{tab:tab1}
\end{table}

%% file: tabs/appendix/00.tex
\begin{table}[t!]
\begin{center}
\begin{small}
\setlength\tabcolsep{3pt}
\scalebox{1}{
\begin{tabular}{lccc}
\toprule
FFHQ 256$\times$256 & \multicolumn{3}{c}{Iterations} \\ 
\arrayrulecolor{gray}\cmidrule(lr){2-4}
{\textit{From-scratch Training}} & {80K} & {90K} & {100K} \\
\arrayrulecolor{gray}\cmidrule(lr){1-4}
DiT-B/2 & {19.18} & {18.80} & {16.55} \\
\textbf{DiT-B/2 + DMP} & {\textbf{15.96}} & {\textbf{15.21}} & {\textbf{13.87}} \\
\arrayrulecolor{black}\bottomrule
\end{tabular}
}
\vspace{\abovetabcapmargin}
\caption{\textbf{Applying DMP from the initial training phase.}
}
\vspace{-4mm}
\vspace{\belowtabcapmargin}
\label{tab:from_scratch}
\end{small}
\end{center}
\end{table}

%% file: tabs/appendix/02.tex
\begin{table}[t!]
\centering
\begin{tabular}{lc}
\toprule
Resolution (256$\times$256) & FFHQ \\
\arrayrulecolor{gray}\cmidrule(lr){2-2}
{Method} & FID{\down} \\
\arrayrulecolor{black}\midrule
Pre-trained DiT-B/2 & 6.27 \\
\quad + Fine-tuning & 6.57$_{(+0.30)}$ \\
\quad + Prompt-tuning & 6.81$_{(+0.54)}$ \\
\quad + LoRA & 7.11$_{(+0.84)}$ \\
\quad \textbf{+ DMP} & \cellcolor{Gray}\textbf{5.87}$_{(-0.40)}$ \\
\bottomrule
\end{tabular}
\vspace{-2mm}
\caption{\textbf{Comparison of DMP and LoRA} using DiT-B/2 model on FFHQ dataset.}
\vspace{-2mm}
\label{tab:tab4}
\end{table}

%% file: tabs/appendix/03.tex
\begin{table}[t!]
\begin{center}
\setlength\tabcolsep{2pt}
\begin{tabular}{lccc}
\toprule
Resolution ($256\times256$)& \multicolumn{3}{c}{ImageNet} \\
\arrayrulecolor{gray}\cmidrule(lr){2-4}
Model & FID{\down} & Precision{\up} & Recall{\up} \\
\arrayrulecolor{black}\midrule 
\multicolumn{2}{l}{\textit{Pre-trained (iter: 7M)}} \\ 
\quad DiT-XL/2 & 2.29 & 0.83 & 0.57 \\
\arrayrulecolor{gray}\cmidrule(lr){1-4}
\multicolumn{2}{l}{\textit{Further Training (iter: 20K)}} \\ 
\quad + Fine-tuning   & 5.04$_{(+2.75)}$ & 0.72 & 0.61 \\
\quad + Prompt tuning & 2.77$_{(+0.48)}$ & 0.81 & 0.59 \\
\arrayrulecolor{gray}\cmidrule(lr){1-4}
\quad \textbf{+ DMP} & \cellcolor{Gray}\textbf{2.25$_{(-0.04)}$} & \cellcolor{Gray}\textbf{0.83} & \cellcolor{Gray}\textbf{0.57} \\
\arrayrulecolor{black}\bottomrule
\end{tabular}
\vspace{-2mm}
\captionof{table}{
\textbf{DMP on DiT-XL/2 model~\cite{peebles2022scalable}.}
}
\label{tab:ditxl}
\vspace{\belowtabcapmargin}
\end{center}
\vspace{-4mm}
\end{table}

%% file: tabs/appendix/04.tex
\begin{table}[t!]
\begin{center}
\scalebox{1}{
\begin{tabular}{lc}
\toprule
\multicolumn{2}{l}{\bf Class-conditional (ImageNet)} \\
\arrayrulecolor{gray}\cmidrule(lr){1-2}
Model & FID{\down} \\
\toprule
DiT-XL/2 (iter: 7M)  &  2.29 \\
\textbf{+ DMP} (only $\vt$) & \cellcolor{Gray}\textbf{2.25} \\
\textbf{+ DMP} ($\vt$ + $\vc$) & \cellcolor{Gray}\textbf{2.25} \\
\arrayrulecolor{gray}\cmidrule(lr){1-2}
\multicolumn{2}{l}{\bf Text-to-Image (MS-COCO)} \\
\arrayrulecolor{gray}\cmidrule(lr){1-2}
Model & FID{\down} \\
\toprule
DiT-B/2 & 7.33 \\
\textbf{+ DMP} (only $\vt$) & \cellcolor{Gray}\textbf{7.30} \\
\textbf{+ DMP} ($\vt$ + $\vc$) & \cellcolor{Gray}\textbf{7.12} \\
\arrayrulecolor{black}\bottomrule
\end{tabular}
}
\vspace{-2mm}
\captionof{table}{
\textbf{Impact of gating condition in DMP.}
\label{tab:gating_condition}
}
\end{center}
\vspace{-15pt}
\end{table}